\pdfoutput=1

\documentclass[11pt,dvipsnames]{article}

\usepackage{amsmath}
\usepackage{amsfonts}
\usepackage{numprint}

\usepackage{graphicx}
\usepackage{tikz}

\usepackage{mathdots}
\usepackage{cancel}
\usepackage{array}
\usepackage{multirow}
\usepackage{amssymb}
\usepackage{gensymb}
\usepackage{tabularx}
\usepackage{booktabs}
\usepackage{caption}
\usepackage{subcaption}
\usepackage{makecell}

\usepackage{algorithm}
\usepackage{algpseudocode}
\usepackage{xcolor}

\usepackage{physics}
\usepackage{amsmath}
\usepackage{yhmath}
\usepackage{siunitx}
\usepackage{extarrows}
\usepackage{csquotes}
\usepackage{mathtools}

\usepackage{acl}

\usepackage{times}
\usepackage{latexsym}

\mathtoolsset{showonlyrefs,showmanualtags}

\usepackage[T1]{fontenc}

\usepackage[utf8]{inputenc}

\usepackage{microtype}

\DeclareMathOperator*{\argmax}{argmax}

\newcommand{\blue}[1]{\textcolor{blue}{#1}}
\newcommand{\red}[1]{\textcolor{red}{#1}}

\newenvironment{itemizesquish}[2]{\begin{list}{\labelitemi}{\setlength{\itemsep}{#1}\setlength{\labelwidth}{#2}\setlength{\leftmargin}{\labelwidth}\addtolength{\leftmargin}{\labelsep}}}{\end{list}}

\title{ `Keep it Together': Enforcing Cohesion in Extractive Summaries \\ by Simulating Human Memory }

\author{Ronald Cardenas\\
  University of Edinburgh \\
  \texttt{ronald.cardenas@ed.ac.uk} \\\And
  Matthias Gall\'e \\
  Cohere \\
  \texttt{matthias@cohere.com}  \\\AND
  Shay B. Cohen \\
  University of Edinburgh \\
  \texttt{scohen@inf.ed.ac.uk} \\}

\begin{document}
\maketitle
\begin{abstract}

Extractive summaries are usually presented as lists of sentences with no expected cohesion between them.
In this paper, we aim to enforce cohesion whilst controlling for informativeness and redundancy in summaries,
in cases where the input exhibits high redundancy.
The pipeline controls for redundancy in long inputs as it is consumed, and balances informativeness and cohesion during sentence selection.
Our sentence selector simulates human memory to keep track of topics --modeled as lexical chains--, enforcing cohesive ties between noun phrases.
Across a variety of domains, our experiments revealed that it is possible to extract highly cohesive summaries that nevertheless read as informative to humans as summaries extracted by only accounting for informativeness or redundancy.
The extracted summaries exhibit smooth topic transitions between sentences as signaled by lexical chains, with
chains spanning adjacent or near-adjacent sentences.

\end{abstract}

\section{Introduction}

Text summarization is the task of processing a document(s) and
producing a shorter text, the \textit{summary}, that retains the gist of the information \cite{nenkova2011automatic}.
\textit{Extractive} summarization selects content units (usually sentences) and presents their concatenation as the summary.
It remains challenging to select the appropriate content units so that the summary ends up non-redundant and informative, 
with much previous work modeling these qualities during document understanding \cite{peyrard-etal-2017-learning,xiao2020systematically,gu2022memsum}.
Moreover, the modeling of summary coherence previously relied on capturing discourse patterns in nearby sentences \cite{barzilay2008modeling,steen-markert-2022-find,zhao2023discoscore}.
Cohesion, a special case of local coherence, relies on the explicit textualization of contextual connections called \textit{cohesive ties},
making a text read as a unified whole \cite{halliday1976cohesion}.

We introduce an extractive summarization methodology that implements two control mechanisms at different stages of processing: the first one to control redundancy during input understanding, and the second one to control the trade-off between informativeness and cohesion during summary extraction.
When building extractive summaries by concatenating sentences, we argue that controlling for cohesion is a better-defined task than aiming to control coherence,
especially if no sort of post-editing (e.g. replacing discourse markers) is applied \cite{zajic2007multi,west2019bottlesum,mallinson2020felix}.
A potential benefit of producing a more cohesive text is that it is easier to read and understand for humans, especially when the knowledge domain is highly technical, as reported by previous work in psycholinguistics \cite{kintsch1990macroprocesses} and automatic summarization \cite{barzilay2002inferring}.

In our pipeline, summary properties are controlled in the following way.
On the one hand, summary redundancy is addressed by controlling the redundancy levels of the input text, following previous findings \cite{carbonell1998use,xiao2020systematically}.
The pipeline consumes input text in a cascaded way: first splitting the input into contiguous passages, then consuming passages one at a time
 so as to minimize their semantic similarity with already selected passages.

On the other hand, informativeness and cohesion are directly modeled during summary extraction.
Extraction is done in a sentence-by-sentence fashion,
quantifying summary properties independently at each step.
The objective is to select a highly cohesive sentence that is informative enough.
We introduce a sentence selector that incrementally builds cohesive chains of noun phrases and models chain interaction.
The selector, \textsc{KvD-Select}, keeps track of chains currently active by simulating human memory according to the Micro-Macro theory, henceforth KvD \cite{kintsch1978toward}, a psycholinguistic theory of discourse comprehension and production.
Working memory --a type of short-term memory-- is modeled as a limited-capacity buffer of lexical chains, forcing the model to keep only the most salient chains. 

We test our methodology on newswire multi-document summarization and single-long document summarization of scientific articles, patents, and government reports.
Across domains, extensive experiments show that, first, our system is effective at incrementally building an input sequence with lower content redundancy, which translated to a significant reduction in summary redundancy.
Second, the proposed sentence selector managed to maintain summaries informative while improving cohesion significantly: over 15\% more noun phrases and over 20\% more sentences were connected through cohesive ties w.r.t a greedy selector.
Tailored human evaluation campaigns revealed that cohesion has a positive impact on perceived informativeness,
and that our extracted summaries exhibit chains covering adjacent or near-adjacent sentences.
Closer inspection showed that topics flow smoothly across extracted summaries with no abrupt change or jumps.


In summary, our contributions are as follows:
\begin{itemizesquish}{-0.3em}{0.5em}
    \item We propose a cascaded encoder capable of consuming arbitrary long textual input that controls the level of content redundancy the rest of the pipeline is exposed to.
    \item We propose a summary extraction method that models informativeness and cohesion independently and allows to control the balance between the two when building the summary.
    \item Automatic and human experiments show the effectiveness of our control mechanisms and how summary properties can be balanced according to user needs in a straightforward way.
\end{itemizesquish}


\section{Related Work}


Previous work modeled cohesion by keeping track of named entities \cite{barzilay2008modeling,guinaudeau2013graph},
topic flow \cite{barzilay2002inferring}, or by implementing discourse theories \cite{jeon2020centering}.
Most similar to our approach, \citet{fang2019proposition} introduced an KvD implementation that models cohesion and informativeness jointly during reading, assigns a single importance score to each sentence, and employs a greedy sentence selector.
In contrast, we quantify summary properties separately, and model cohesion by
implementing KvD during sentence selection.
This approach allows more explicit control of the contribution of each property during summary extraction.

Similar ways to control summary properties during selection have focused only on minimizing redundancy \cite{carbonell1998use,fabbri2019multi,xiao2020systematically}, where the extractive summary is regarded as a list of sentences with no particular order to them, a design choice possibly influenced by the format of available benchmarks such as CNN/DM \cite{hermann2015teaching}.
However, seminal work highlighted the role of redundancy \cite{walker1993informational,tauste1995comodidad}, and how its presence is a result of human memory limitations \cite{johnstone1994repetition}.

In this work, we provide evidence that controlling for cohesion constitutes a better strategy for providing the end-user with a more comprehensible summary, formatted as a multi-sentence cohesive text.
Our results show that this setup is especially effective when the knowledge domain is highly technical, and when a sentence ordering cannot be inferred from the input trivially, e.g. in multi-document summarization.

Previous work on passage selection focused on selecting salient passages (paragraphs in the a document or documents in a cluster) from long and noisy input by ranking them according to TF-IDF similarity with the article title \cite{liu2018generating} or with a learned ranker \cite{liu2019hierarchical}.
In contrast, our redundancy control method operates in an unsupervised fashion by selecting passages according to their relevance, measured by LexRank~\cite{erkan2004lexrank}, and to their redundancy w.r.t.\ already selected passages.




\section{Problem Formulation}

\begin{figure}[t]
    \centering
    \includegraphics[width=0.4\textwidth]{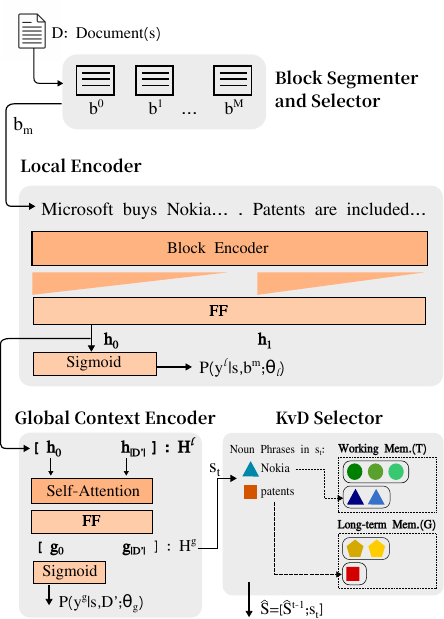}
    \caption{Our extraction pipeline:
    local extraction step $m$ adds local sentences to $D'$; at sentence selection step $t$, KvD-Select balances informativeness of candidate $s_t$ with cohesion of summary $\hat{S}$. }
    \label{fig:architecture}
\end{figure}

We tackle the task of extractive summarization as a sentence scoring step followed by a selection step.
Figure~\ref{fig:architecture} shows the pipeline of the system, in which sentences are scored in a cascaded fashion, as follows.
First, the input is segmented into blocks of contiguous sentences to be selected based on their relevancy and their redundancy w.r.t.\ already selected blocks.
Then, a local encoder obtains block-level representations for each sentence in the block.
After all document blocks are processed, the encodings are concatenated into a single embedding sequence and passed to the global context encoder, which will obtain a
document-aware representation of each sentence.
Finally, a selection module will extract a subset of sentences 
and present them as the summary in the order they were extracted.
We now proceed to elaborate on each module of the proposed pipeline.

\subsection{Block Segmentation and Selection}
\label{section:block-sel}
Processing starts by segmenting the document(s) $D$
into fixed-length overlapping blocks, each of which includes preceding and subsequent wordpieces, providing surrounding context.
Then, blocks are selected iteratively until a predefined budget (total number of wordpieces) is met.
At step $m$, block $b_m$ is selected such that
\vspace{-0.1cm}
\begin{equation}
\label{eq:block-sel}
\small
b_m = \argmax_{b \in B \setminus \hat{B} } [ \lambda_{b}\text{LR}(b) - (1-\lambda_{b}) \max_{b_j \in \hat{B}} \text{Sim}(b,b_j) ]
\end{equation}
\noindent where $\hat{B}$ is the set of blocks already selected, $\text{Sim}(x,y)$ is the cosine similarity between TF-IDF vectors of blocks $x$ and $y$, and $\lambda_b$ allows to control the mix of both terms.
$\text{LR}(b)$ is the continuous LexRank score of block $b$ \cite{erkan2004lexrank},
calculated over the complete graph of blocks in $D$,
\vspace{-0.1cm}
\begin{equation}
\small
    \text{LR}(b) = \frac{d}{|B|} + (1-d) \sum_{v \in \text{adj}[b]} \frac{\text{Sim}(b,v)}{ \sum_{z \in \text{adj}[v]} \text{Sim}(z,v) } \text{LR}(v)
\end{equation}
\noindent where $d$ is the damping factor and $\text{adj}(b)$ is the set of block nodes adjacent to $b$.
This module balances block relevancy (as proxied by centrality) and input redundancy
in a straightforward way by linearly combining their scores.
After an optimal block is selected, it is sent to the local encoder module.

\subsection{Local Encoder (LE)}
Given block $b$ as a sequence of wordpieces spanning contiguous sentences,
the local encoder will obtain representations for each sentence covered in $b$.
This module is trained as a local extractive summarizer itself, 
under sequence labeling formulation where each
sentence in the block is labeled as $y^{\ell}_i \in \{0,1\}$ to indicate whether sentence $s_i$ is selected or not.
Then, sentence representation $h_i$ is defined as the average embedding over $s_i$ wordpieces, obtained from a LongT5 encoder \cite{guo2022longt5}.
Finally, the probability of $s_i$ being locally selected is defined as $P(y^{\ell}_i \mid s_i,b;\theta_{\ell})= \sigma( W^{\ell} \cdot h_i )$,
and the module is trained using cross-entropy loss independently from the rest of the pipeline.
During inference, the local encoder consumes one block, selects $N$ sentences and adds them to $D'$ --containing all locally selected sentences so far--, and their
corresponding embeddings to  $H^{\ell}$.

\subsection{Global Context Encoder (GCE)}
Given the sequence of local sentence embeddings $H^\ell$,
this module obtains the sequence of globally-aware representations $H^g$ as follows.
Sequence $H^\ell$ is passed through a self-attention layer \cite{vaswani2017attention},
i.e.\ $g_t=\text{SelfAttn}(h_t,H^{\ell}), \forall h_t \in H^{\ell}$.
Similarly to the LE module, the probability of selecting $s_t \in D'$ is $P(y^g_t \mid s_t,D';\theta_g)= \sigma( W^g \cdot g_t )$,
where $y^g_t \in \{0,1 \}$ indicates whether $s_t$ is selected or not for the final, global summary,
and also trained using cross-entropy loss.

\subsection{Sentence Selection}
Finally, candidate summary $\hat{S}$ is built by selecting one sentence at a time from $D'$, 
taking into account the informativeness and cohesion of each candidate sentence 
w.r.t.\ the already selected sentences.
At selection step $t$, the optimal sentence is given by
\vspace{-0.1cm}
\begin{equation}
\label{eq:emb-trade}
    s_t = \argmax_{s \in D' \setminus \hat{S}^{t-1}} [\lambda_{\text{sel}} f_{\text{I}}(s) + (1-\lambda_{\text{sel}}) f_{\text{C}}(\hat{S}^t)]    
\end{equation}
\noindent where function $f_{\text{I}}$ estimates the informativeness of candidate sentence $s$, 
$f_{\text{C}}$ estimates the cohesion of candidate summary $\hat{S}^t=[\hat{S}^{t-1};s]$, and $\lambda_{\text{sel}} \in [0,1]$ is a parameter that allows to control their trade-off.
Following \newcite{xiao2020systematically}, we take the probability of selecting $s$ given by the global context encoder module as a proxy for informativeness, i.e.\
$f_{\text{I}}(s)=P(y^{g} \mid s,D';\theta_g)$.
In the next section, we elaborate on how $f_{\text{C}}$ models and enforces cohesion during sentence selection.


\section{Cohesion during Summary Extraction}

Cohesion is a language mechanism that enables a sequence of sentences to function as a unified whole \cite{halliday1976cohesion}.
It does so by linking semantic units in a text through \textit{cohesive ties},
regardless of the grammatical or discourse structure these units are part of.
In particular, lexical cohesion links units with the same lexical form, synonyms, or units in the same semantic field.
Furthermore, units tied cohesively can be grouped in chains by their semantic similarity.
Whilst the mere presence of two or more chains does not guarantee a cohesive effect, their interaction can be a reliable proxy for cohesion \cite{morris1991lexical,barzilay-elhadad-1997-using}.

In this paper, we focus on modeling lexical cohesive ties between noun phrases in nearby sentences of a summary by controlling the interaction between lexical chains.

\subsection{KvD Select}

The proposed selector, \textsc{KvD-Select}, calculates cohesion score $f_{C}$ by simulating the processes in working memory during text production.
The procedure is based on the Micro-Macro Structure theory \cite{kintsch1978toward}, which describes the cognitive processes involved in text comprehension and production at the local (micro) and global (macro) level of discourse.
Following \newcite{fang2019proposition}, we implement processes happening at micro-level, which deal with the movement of content in and out of working memory.

Let $T$ be working memory and $G$ long-term memory (LTM), where both are separate sets of cohesive chains, and each chain as a set of noun phrases (NPs).
At selection step $t$, the algorithm extracts NPs from $s_t$ and connects them to the chains in $T$ and $G$, constraining the number of active chains in $T$ afterward.
Cohesive score $f_{\text{C}}$ then depends on the average similarity between units added to $T$ and those added to $G$.
We now elaborate on each step of the algorithm.

\paragraph{Extracting Noun Phrases.}
Given sentence $s_t \in D'$, we obtain $P$, the set of extracted nominal chunks,
obtained by merging nominal nodes in dependency trees with their children, following the procedure of \citet{fang2019proposition}.
Specifically, given that node \textit{u} is nominal dependent of a clausal predicate,
\textit{u} will have its child \textit{v} merged if either\textit{v} is a function word, a single-token modifier, or
\textit{u} and \textit{v} form part of a multi-word expression.

\paragraph{Adding Content to Memory.}
Next, cohesive ties between $s_t$ and $\hat{S}^{t-1}$ are enforced by adding
each NP in $P$ to the chain with the highest element-wise semantic similarity.
Formally, the optimal chain to add $a \in P$ to is
$C^* = \argmax_{C \in T} \{ \phi(p,C) \}$,
where $\phi$ is the average BERTScore \cite{zhang2019bertscore} between $a$ and each NP in $C$.
In order to make sure that chains maintain an acceptable level of semantic similarity between elements, $a$ is added to chain $C$ only if
$\phi(a,C) \geq \nu$, where $\nu$ is the minimum admissible similarity.
This way the algorithm can control the similarity length between chain members, and avoid a single, long chain.
%

If similarity with chains in $T$ is not strong enough, we look at chains in $G$, in which case the chosen chain is moved back to $T$.
This step simulates how humans recall content no longer present in WM, the \textit{recall mechanism} \cite{kintsch1978toward}. If still no chain in $G$ meets the similarity requirement, we proceed to create a brand new chain in $T$ with $a$ as its sole element.
By searching for a good enough candidate chain first in $T$ and then in $G$,
we encourage cohesive ties between NPs in nearby sentences.

\paragraph{Updating Memory.}
After adding incoming NPs to chains in memory, $T$ is updated to retain only the \texttt{WM} most recent chains,
where \textit{recency} of a chain is defined as the id of the selection step in which this chain was last retained in $T$.
For instance, a chain currently in $T$ is more recent (higher step id) than a chain in $G$ discarded in an earlier step.
This design choice mimics the \textit{recency effect} behaviour during \textit{free recall} tasks in human subjects \cite{glanzer1972storage}, a behaviour attributed to short-term memory.
Finally, discarded chains are moved to $G$, concluding the selection step.

\paragraph{Candidate Scoring.}
Next, we define cohesion score $f_{\text{coh}}$ which will be used to discriminate amongst possible continuations to $\hat{S}^{t-1}$.
The objective is to encourage NPs in $P$ to be assigned to recent chains, in turn encouraging chains to cover nearby sentences in the final summary.
In addition, we want to score down candidate sentences with NPs added to chains in long-term memory.

Let $A_T=\{a; a \in P, C_a \in T \}$, where $C_a$ is the chain $a$ was added to.
Similarly, let $A_G=\{b; b \in P, C_b \in G \}$.
Then, let $\text{rec}(C)$ be the number of selection steps passed since the last time chain $C$ was retained in $T$.
Quantity $\text{rec}(C)$ functions as a proxy for how spread chain $C$ is, i.e.\ how far away two sentences covered by $C$ are.
Then,
\vspace{-0.1cm}
\begin{align}
    f_{\text{coh}} = \frac{1}{|A_T|} \sum_{a \in A_T} \frac{\phi(a,C_a)}{\text{rec}(C_a)} + \frac{\gamma_{\text{rec}}}{|A_G|} \sum_{b \in A_G} \frac{\phi(b,C_b)}{\text{rec}(C_b)}.
\end{align}
Hence, the cohesive score depends on the contribution of each cohesive tie formed.
For each chunk in $A_T$ and $A_G$, its contribution depends directly on the strength of similarity to its assigned chain
and inversely on the spread of said chain.
The contribution of chunks in $A_G$ is scaled down by hyper-parameter $\gamma_{\text{rec}} \in [ 0;1 ]$ as to simulate the higher cognitive cost
incurred when retrieving information from long-term memory.




    

\section{Experimental Setup}

We now describe the datasets used, training details, baselines, and evaluation methodology.

\subsection{Datasets}
We employ datasets pairing single and multi-documents with human-written summaries,
specifically scientific articles from the biomedical domain (PubMed; \citet{cohan2018discourse})\footnote{We use text from all sections as the source document and the abstract as reference summary.}, patents in the Chemistry and Metallurgy industry (BigPatent.C;\citet{sharma2019bigpatent}), legislature reports of U.S. bills (GovReport; \citet{huang2021efficient}), and news articles (MultiNews;\citet{fabbri2019multi}).

\subsection{Pipeline Parameters}

Hyper-parameters were tuned over the validation sets of each dataset.

\paragraph{Document Segmentation and Block Selection.}
We use a block size of $B=2048$, context size of $C=200$ pieces, $\lambda_b=0.2$,
and set a budget of \num{16384} input wordpieces. 


\paragraph{Local Encoder (LE), Global Context Encoder (GCE).}
The block encoder in LE is initialized with a pretrained checkpoint of LongT5 with transient-global attention \cite{guo2022longt5},\footnote{HuggingFace, \texttt{google/long-t5-tglobal-base}} and an output layer of size \num{200}.

The LE module is trained independently from the GCE module,
with LE being trained first, then GCE trained whilst LE remains frozen.
In both cases, we used the Adam optimizer \citep{loshchilov2018decoupled}, a constant learning rate of $1e^{-6}$, effective batch size of $64$, and  \num{50}k training steps.
During inference, we extract a maximum of $N=10$ local sentences per block
and a maximum of \num{1000} sentences in total.

\paragraph{Summary Extractor.}
We set $\lambda_{sel}=0.8$, working memory \texttt{WM}$=6$, recall cost $\gamma_{\text{rec}}=0.01$, and a minimum NP similarity of $\nu=0.6$.
Word budget is set to \num{200}, \num{100}, \num{650}, \num{250} for PubMed, BigPatent.C, GovReport, and MultiNews, respectively.


\subsection{Comparison Systems}
\label{sec:base-sels}
Baselines consist of encoders pre-trained to consume the whole input as a sequence of up to \num{16384} pieces long, topped with a sigmoid layer for classification.
As encoders, we employ the base pretrained checkpoints of Longformer (LED-Ext;\citet{Beltagy2020Longformer}) and LongT5's encoder (LT5-Ext;\citet{guo2022longt5}).\footnote{Similar model such as BigBird \cite{zaheer2020big} and Linformer \cite{wang2020linformer} do not have the same input capacity, hence they were not considered. However, we do include other options of Local Encoders in \S~\ref{apx-additional-systems}. }

The impact of cohesion modeling is assessed by employing a greedy selector over GCE scores, equivalent to set $f_\text{C}=0$ in Eq.~\refeq{eq:emb-trade}, dubbed \textsc{BlockSel[LT5]}.
Finally, we compare against the following (summary) property-oriented selectors.

\paragraph{MMR-Select.} \cite{xiao2020systematically}
Reduces redundancy by
selecting $s_i$ (candidate sentence at selection step $i$) such that cosine similarity w.r.t.\ the partially extracted summary $\hat{S}$ is minimized. Informativeness and redundancy are balanced in the same way as in Eq.~\refeq{eq:emb-trade}.

\paragraph{N-gram passing (NPass).} 
Encourages repetition by allowing $p$ percent of n-grams in $s_i$ to overlap with $\hat{S}$.
When $p=0$, this method reduces to n-gram blocking, whereas when $p=1.0$, to greedy selection.
We report bi-gram passing with $p=0.8$.



\paragraph{Shuffle Classifier (CCL-Select).}
Holistically quantifies local coherence using CCL \cite{steen-markert-2022-find}, a scorer trained to distinguish shuffled from unshuffled text.
We use RoBERTa \cite{roberta-model} as underlying model and use a window of \num{3} consecutive sentences.



\subsection{Evaluation}

Informativeness is assessed using ROUGE F$_1$ \cite{lin2004rouge},
Redundancy is evaluated using sentence-wise ROUGE-L (RdRL;\citet{bommasani2020intrinsic}), and inverse uniqueness (IUniq;\citet{peyrard-etal-2017-learning}), defined as $1 - \text{Uniqueness}$, where `Uniqueness' 
is the ratio of unique n-grams to the total number of n-grams.
We report the mean value between uni-, bi-, and trigrams.
Higher values denote higher redundancy.

Cohesion is measured using Entity Graph (EGr;\citet{guinaudeau2013graph}), which models a text as a sentence graph with edges between sentences with nouns in common, using the average edge weight as a proxy for cohesion. 
Finally, local coherence is assessed using CCL \cite{steen-markert-2022-find}.

\subsubsection{Human Evaluation}
We elicit human judgments to assess overall quality, informativeness, and cohesion in two separate studies.
We sampled 30 documents from the test set of \textsc{PubMed} and compare systems \textsc{BlockSel[LT5]}, 
\textsc{MMR-Select}, and \textsc{KvD-Select}.

\paragraph{Ranking Campaign.}
Subjects were shown the abstract and the introduction of a scientific article along with two system summaries, and then
then asked to select the best summary (or select both in case of tie) according to three criteria:
(i) overall quality, (ii) informativeness, and (iii) cohesion.
In this setup, cohesion is evaluated as a holistic property of the text, as perceived by a reader.

\paragraph{Chaining Campaign.}
Subjects were shown a single summary and were asked to annotate lexical chains by grouping together pre-extracted NPs in the same semantic field.
We report the following chain properties:
(i) \textit{chain spread}, defined as the average number of sentences between immediate-neighbor sentences covered by the same chain;
(ii) \textit{chain density}, the number of chains covering the same sentence;
and (iii) \textit{sentence coverage}, the percentage of sentences covered by at least one chain.

Inter-annotator agreement is calculated as the average lexical overlap between chains, expressed in F$_1$ score,
calculated pair-wise between subjects.
For this study, we include reference summaries as one more analysis system.


\section{Results and Discussion}

Next, we discuss the results of our analyses and the outcome of the human evaluation campaigns.

\subsection{Reducing Redundancy in Input Blocks}
The following block selection strategies were compared:
(i) \textit{Original}, consisting of selecting blocks in their original order in the source document;\footnote{For MultiNews, we use the order provided in the dataset.}
(ii) \textit{Oracle Selection}, which selects the block that maximizes ROUGE F$_1$ scores (mean of ROUGE-1 and ROUGE-2) w.r.t.\ the reference summary;
(iii) \textit{Max. Redundancy}, which selects the most similar block possible (by flipping the sign in Eq.~\refeq{eq:block-sel});
and finally, (iv) \textsc{BlockSelect}, the proposed strategy.

\begin{figure*}[ht]
    \centering
    \includegraphics[width=0.85\textwidth]{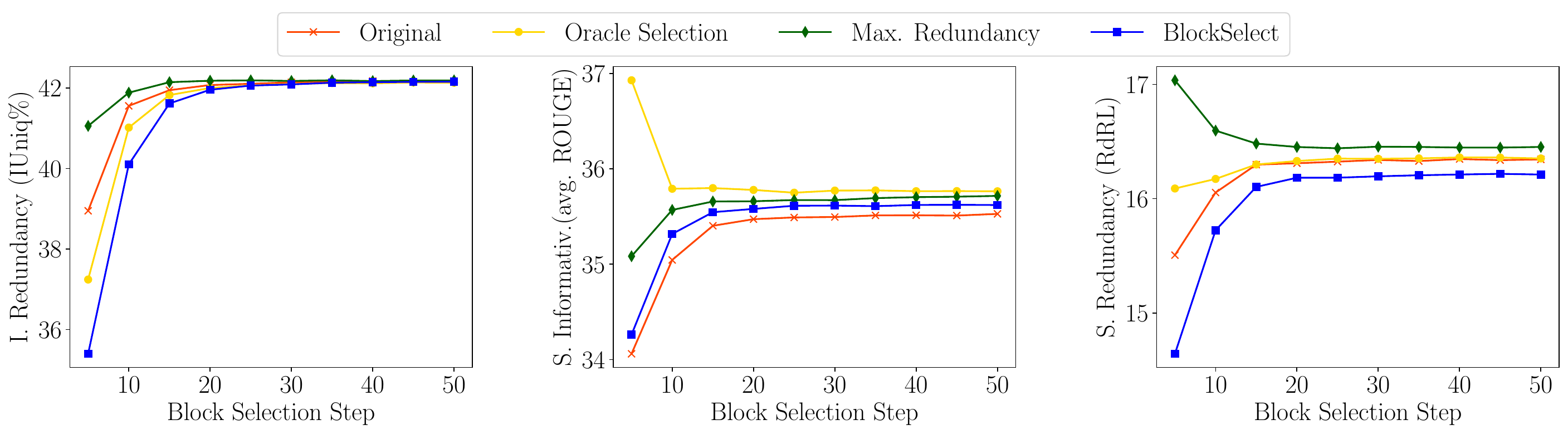}
    \caption{Effect of block selection strategy over input redundancy (left), summary informativeness (center), and summary redundancy (right), evaluated as block selection proceeds on the \textsc{MultiNews} validation dataset.}
    \label{fig:brank}
\end{figure*}

The analysis, showcased in Figure~\ref{fig:brank}, evaluates input redundancy at each block selection step,
as well as informativeness and redundancy of summaries extracted from the blocks available at each step, using a greedy selector.
We observe that \textsc{BlockSelect} has a direct impact not only on input redundancy but also on summary redundancy.
As intended, it is effective at incrementally building an input sequence with lower RdRL, and the extracted summaries exhibit lower RdRL and comparable ROUGE scores than the compared block selection strategies.
Similar trends were observed in the other datasets.\footnote{See Fig.~\ref{fig:brank-all} in \S~\ref{apx-brank-all} for results in other datasets.}

\subsection{Trading-off Informativeness and Cohesion}
Next, we turn to the summary extraction module.
Tables~\ref{table:inf-res-1} and \ref{table:inf-res-2} present the performance in terms of informativeness, whereas Table~\ref{table:coh-res}, in cohesion.
In all our experiments, statistical significance at the 95\% confidence level is estimated using Mann–Whitney U tests ($p<0.05$).


First, note the impact on cohesion when controlling for redundancy.
MMR-Select manages to obtain comparable informativeness levels to BlockSel[LT5], being most effective for \textsc{BigPatent.C}.
However, minimizing sentence similarity comes at the expense of a significant decrease in cohesion (EGr) and coherence (CCL).
Second, we find that \textsc{NPass} is the only one capable of obtaining comparable or better ROUGE scores
but EGr scores indicate that lexical passing is not enough to improve cohesion.





When guiding selection with a holistic shuffle scorer, as expected, \textsc{CCL-Select} obtains remarkably high CCL scores.
However, note that this selector does show a significant reduction in EGr scores w.r.t.\ BlockSel[LT5],
indicating that CCL is measuring also discourse organization, possibly in the form of rhetorical role ordering --first background, then method, and so on.
Hence, it can be said that summaries in \textsc{CCL-Select} are better organized in terms of rhetorical roles but exhibit lower cohesion than greedily selected summaries.


Finally, \textsc{KvD-Select} manages to strike an even more aggressive trade-off between informativeness and cohesion.
Across datasets, the selector exhibits lower ROUGE scores but the best  EGr scores (except for \textsc{PubMed}), and second highest CCL score after \textsc{CCL-Select}.


\begingroup

\setlength{\tabcolsep}{4pt} 

\begin{table}[ht]
\centering
\scriptsize
\begin{tabular}{lllllll}
\toprule
\multicolumn{1}{c}{\multirow{2}{*}{\textbf{System}}} & \multicolumn{3}{c}{\textbf{PubMed}}                                                                 & \multicolumn{3}{c}{\textbf{BigPatent.C}}    \\
\multicolumn{1}{c}{}                                 & \multicolumn{1}{c}{\textbf{R1}} & \multicolumn{1}{c}{\textbf{R2}} & \multicolumn{1}{c}{\textbf{RL}} & \multicolumn{1}{c}{\textbf{R1}} & \multicolumn{1}{c}{\textbf{R2}} & \multicolumn{1}{c}{\textbf{RL}} \\ \toprule
LED-Ext                                             & 40.20                           & 13.87                           & 36.85                           & 36.65                           & 11.07                           & 31.94                           \\
LT5-Ext                                          & \textbf{48.15}                           & \textbf{21.45}                           & \textbf{44.49}                           & 39.54                           & 13.25                           & \textbf{34.30}                           \\
BlockSel[LT5]                                                &        46.16\dag                         &       19.74\dag                          &       42.49\dag                          & 39.57\dag                           & 13.25\dag                           & 34.26\dag                            \\ \midrule
+MMR-Select                                           &       \red{46.14}\dag                          &       \red{19.63}\dag                          &      \blue{42.47}\dag                           & \blue{\textbf{39.59}}\dag                           & \blue{\textbf{13.29}}\dag                           & \blue{\textbf{34.30}}\dag                                                      \\
+NPass                                             &         \blue{46.38}\dag                        &       \blue{19.92}\dag                          &        \blue{42.74}\dag                         & \blue{\textbf{39.59}}\dag                           & \blue{13.26}\dag                           & \blue{34.29}\dag                            \\
+CCL-Select                                           &   \red{45.91}\dag                              &       \red{19.60}\dag                          &      \red{42.45}\dag                           & \red{39.16}\dag                           & \red{12.95}                           & \red{33.92}\dag   \\
+\textbf{KvD-Select}                                           &   \red{44.90}                              &       \red{18.47}                          &      \red{41.27}                           & \red{38.37}                           & \red{12.41}                           & \red{33.13}                                                 \\ \bottomrule
\end{tabular}
\caption{Informativeness in terms of ROUGE scores (R1, R2, RL) for PubMed and BigPatent.C datasets.
\dag: no stat.\ difference between systems in the same column.
Best systems are \textbf{bolded}; systems better than \textsc{BlockSel[LT5]} shown in \blue{blue} and worse, in \red{red}.}
\label{table:inf-res-1}
\end{table}

\endgroup

\begingroup

\setlength{\tabcolsep}{4pt} 

\begin{table}[ht]
\centering
\scriptsize
\begin{tabular}{lllllll}
\toprule
\multicolumn{1}{c}{\multirow{2}{*}{\textbf{System}}} &  \multicolumn{3}{c}{\textbf{GovReport}}                                                              & \multicolumn{3}{c}{\textbf{MultiNews}}                                                              \\
\multicolumn{1}{c}{}                                 & \multicolumn{1}{c}{\textbf{R1}} & \multicolumn{1}{c}{\textbf{R2}} & \multicolumn{1}{c}{\textbf{RL}} & \multicolumn{1}{c}{\textbf{R1}} & \multicolumn{1}{c}{\textbf{R2}} & \multicolumn{1}{c}{\textbf{RL}} \\ \toprule
LED-Ext                                         &    57.59                           & 23.40                           & 54.56                           & 45.28                           & 15.78                           & 41.35                           \\
LT5-Ext                                           & 59.33                           & 25.94                           & 56.29                           & \textbf{47.07}                           & \textbf{17.54}                           & \textbf{42.96}                           \\
BlockSel[LT5]                                                &  59.73\dag                           & 26.21\dag                           & 56.50\dag                           & 46.80\dag                           & 17.21\dag                           & 42.66\dag                           \\ \midrule
+MMR-Select                                           &        \blue{\textbf{59.79}}\dag                           & \blue{\textbf{26.30}}\dag                           & \blue{\textbf{56.56}}\dag                           & \red{46.76}\dag                           & \red{17.13}\dag                           & \red{42.59}\dag                           \\
+NPass                                             &         \blue{\textbf{59.79}}\dag                           & \blue{26.25}\dag                           & \blue{\textbf{56.56}}\dag                           & \blue{46.91}\dag                           & \blue{17.27}\dag                           & \blue{42.78}\dag                           \\
+CCL-Select                                           &  \red{59.72}\dag                           & \blue{26.24}\dag                           & 56.50                           & \blue{46.85}\dag                           & \blue{17.29}\dag                           & \blue{42.71}\dag                           \\
+\textbf{KvD-Select}                                           &   \red{57.88}                           & \red{23.66}                           & \red{54.57}                           & \red{45.85}                           & \red{16.13}                           & \red{41.62}                           \\ \bottomrule
\end{tabular}
\caption{Informativeness in terms of ROUGE scores (R1, R2, RL) for GovReport and MultiNews datasets.
See Table~\ref{table:inf-res-1} for formatting details.}
\label{table:inf-res-2}
\end{table}

\endgroup


\begingroup
\setlength{\tabcolsep}{4pt} 

\begin{table*}[ht]
\centering
\footnotesize
\begin{tabular}{lllllllllllll}
\toprule
\multirow{2}{*}{\textbf{Systems}} & \multicolumn{3}{c}{\textbf{PubMed}}                                                                       & \multicolumn{3}{c}{\textbf{BigPatent.C}}                                                                  & \multicolumn{3}{c}{\textbf{GovReport}}                                                                    & \multicolumn{3}{c}{\textbf{MultiNews}}                                                                    \\
                                  & \multicolumn{1}{c}{\textbf{RdRL}} & \multicolumn{1}{c}{\textbf{EGr}} & \multicolumn{1}{c}{\textbf{CCL}} & \multicolumn{1}{c}{\textbf{RdRL}} & \multicolumn{1}{c}{\textbf{EGr}} & \multicolumn{1}{c}{\textbf{CCL}} & \multicolumn{1}{c}{\textbf{RdRL}} & \multicolumn{1}{c}{\textbf{EGr}} & \multicolumn{1}{c}{\textbf{CCL}} & \multicolumn{1}{c}{\textbf{RdRL}} & \multicolumn{1}{c}{\textbf{EGr}} & \multicolumn{1}{c}{\textbf{CCL}} \\ \midrule
LED-Ext                          & \textbf{14.70}                             & 1.00                               & 0.36                             & \textbf{14.94}                             & 0.69                               & 0.35                             & \textbf{15.06}                             & 1.91                               & 0.30                             & \textbf{11.25}                             & 0.81                               & 0.30                             \\
LT5-Ext       & 16.49                & \textbf{1.10}                     & 0.24                & 19.76                     & 0.75                          & 0.32                     & 15.78                   & 2.01                        & 0.31                   & 12.24                   & 0.91                        & 0.25  \\
BlockSel[LT5]       & 17.08\dag                & 1.07\dag                     & 0.25\dag                & 20.15\dag                     & 0.73                          & 0.37\dag                     & 16.34                   & 2.04\dag                        & 0.30\dag                   & 12.26                   & 0.90\dag                        & 0.24                   \\ \midrule
+MMR-Select    & \blue{16.99}\dag                & 1.07                     & \red{0.23}\dag                & \blue{19.17}\dag                     & 0.73                          & \red{0.33}\dag                     & \blue{16.16}                   & \red{2.03}\dag                        & 0.30                   & \blue{12.05}\dag                   & \red{0.88}\dag                        & \red{0.22}                   \\
+NPass      & \blue{16.39}                & 1.07                     & 0.25                & \blue{19.79}\dag                     & 0.73                          & 0.37\dag                     & \blue{16.24}                   & 2.04                        & \blue{0.31}\dag                   & \blue{12.03}\dag                   & \red{0.89}\dag                        & 0.24                   \\
+CCL-Select    & \blue{16.63}\dag                & \red{1.06}\dag                     & \blue{\textbf{0.66}}                & \blue{18.97}                     & \red{0.71}                          & \blue{\textbf{0.88}}                     & \blue{16.31}                   & 2.04\dag                        & \blue{\textbf{0.65}}                   & \blue{11.93}                   & \red{0.86}                        & \blue{\textbf{0.63}}                   \\
\textbf{+KvD-Select}    & \blue{16.24}\dag                & \red{1.05}\dag                     & \blue{0.27}                & \red{21.09}                     & \blue{\textbf{0.78}}                          & \blue{0.40}\dag                     & \red{16.69}                   & \blue{\textbf{2.15}}                        & \blue{0.32}                   & \red{12.97}                   & \blue{\textbf{0.99}}                        & \blue{0.28}                   \\ \midrule
Gold           & 13.54                & 0.95                     & 0.87                & 18.11                     & 0.78                          & 0.92                     & 13.37                   & 1.95                        & 0.87                   & 9.72                   & 0.71                        & 0.90                   \\
\bottomrule
\end{tabular}
\caption{Summary redundancy (RdRL), cohesion (EntityGraph; EGr), and local coherence (CCL).
For cohesion and coherence metrics, higher is better.
See Table~\ref{table:inf-res-1} for formatting details.
}
\label{table:coh-res}
\end{table*}
\endgroup

\begin{figure*}[ht]
    \centering
    \includegraphics[width=0.8\textwidth]{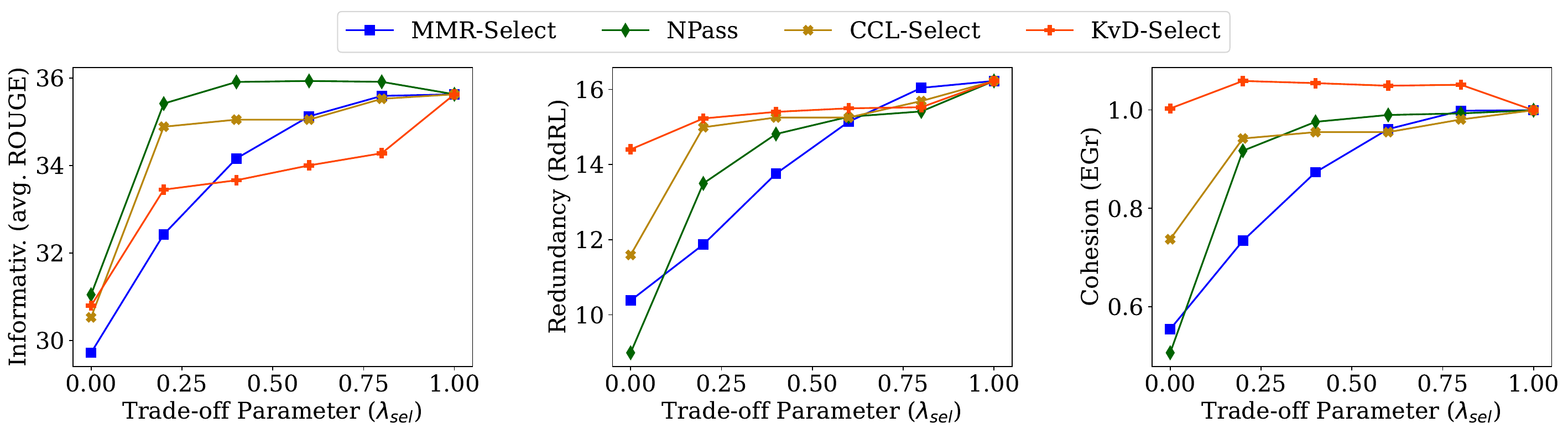}
    \caption{Informativeness (right), redundancy (center), and cohesion (right) in summaries, across increasing values of trade-off parameter $\lambda_{\text{sel}}$, on the validation set of \textsc{MultiNews}.}
    \label{fig:lambsel-vals}
\end{figure*}

\paragraph{Effect of Parameter $\lambda_\text{sel}$.}
Figure~\ref{fig:lambsel-vals} showcases how summary properties vary across increasing levels of $\lambda_\text{sel}$, for \textsc{MultiNews}.
Note that when $\lambda_\text{sel}=0$ selectors depend entirely on $f_{\text{C}}$,
and $\lambda_\text{sel}=1.0$ is the greedy selector.
As expected, informativeness is higher as $f_{\text{I}}$ is weighted up (higher $\lambda_\text{sel}$) with all selectors except \textsc{MMR-Select}. This indicates that it is possible to increase cohesion whilst preventing redundancy from increasing, however at the expense of decreased informativeness.
Interestingly, \textsc{KvD-Select} seems robust to $\lambda_\text{sel}$ in terms of EGr and RdRL.
We hypothesize that \textsc{KvD-Select} benefits from a signal indicating which cohesive ties are informative and worth enforcing.

\paragraph{Impact of Block Selection.}
Systems using our block selection mechanism exhibit lower ROUGE scores than flat processing baselines (LED-Ext and LT5-Ext) in \textsc{PubMed} and \textsc{MultiNews}, and
comparable performance for \textsc{BigPatent.C} and \textsc{GovReport}.
However, BlockSel[LT5] shows slightly higher EGr scores in all datasets. 
This indicates that cascaded processing puts a greedy selector in a better position to extract 
more cohesive summaries at the expense of a slight decrease in informativeness.


\subsection{Human Evaluation}

In both studies, statistical significance between system scores was assessed using a one-way ANOVA with posthoc Tukey tests with $95\%$ confidence interval ($p<0.01$).
Results are presented in Table~\ref{c3-table:human-eval-res}.

\paragraph{Ranking.}
Krippendorff’s $\alpha$ \cite{krippendorff2011computing} showed an inter-annotator agreement of $0.68$.
For overall quality, subjects showed a significant preference for \textsc{KvD-Select} over BlockSel[LT5].
For cohesion, KvD-Selec was perceived as more cohesive compared to BlockSel[LT5], and
BlockSel[LT5] was more cohesive than MMR-Select.

\paragraph{Chaining.}
Chain overlap was calculated at $0.90$.
Differences between BlockSel[LT5] and all other systems, as well as 
MMR-Select--Gold and KvD-Select--BlockSel[LT5] were found to be significant, 
for all measurements of cohesion.
Moreover, the number of NPs annotated per chain was \num{2.30}, \num{2.33}, \num{2.80}, and \num{2.55},
for systems BlockSel[LT5], MMR-Select, KvD-Select, and Gold, respectively.

We found that KvD-Select summaries exhibit more active and denser chains and better-covered sentences than the baselines.
Note that BlockSel[LT5] obtains the lowest chain spread but also low coverage,
indicating that its summaries exhibit very few chains that happen to be close to each other.
In contrast, MMR-Select obtains the highest chain spread and low number of chains,
indicating content with low diversity and sparsely presented.

\begingroup
\setlength{\tabcolsep}{4pt} 

\begin{table}[h]
\centering
\scriptsize
\begin{tabular}{lccc|ccc}
\toprule
\multicolumn{1}{c}{\multirow{2}{*}{\textbf{System}}} & \multicolumn{3}{c}{\textbf{Ranking}}                                                                          & \multicolumn{3}{c}{\textbf{Chaining}}                                                                                                                                                                \\
\multicolumn{1}{c}{}                                 & \multicolumn{1}{c}{\textbf{Ov}$\downarrow$} & \multicolumn{1}{c}{\textbf{I}$\downarrow$} & \multicolumn{1}{c|}{\textbf{C}$\downarrow$} & \multicolumn{1}{l}{\textbf{Spr}$\downarrow$} & \multicolumn{1}{l}{\textbf{Den}$\uparrow$} & \multicolumn{1}{l}{\textbf{Cov$\uparrow$}}  \\ \midrule
BlockSel[LT5]                                               & 1.59                                 & 1.56                              & 1.59                               & \textbf{1.93}                                  & 1.29                                 & 57.12                                   \\
+MMR-Select                                           & 1.50                                 & 1.48                              & 1.47                               & 2.36                                  & 1.28                                 & 53.21                                   \\
+\textbf{KvD-Select}              & \textbf{1.41}                                 & \textbf{1.46}                              & \textbf{1.44}                               & 2.05                                  & \textbf{1.40}                                 & \textbf{68.78}                             \\ \midrule
Gold                                                 & -                                    & -                                 & -                                  & 1.91                                  & 1.36                                 & 69.65 \\ \bottomrule
\end{tabular}
\caption{Ranking (left) w.r.t.\ (Ov)erall quality, (I)nformativeness, and (C)ohesion;
and properties of annotated chains (right): 
spread (Spr), density (Den), and sentence coverage (Cov,\%).
Best systems are \textbf{bolded}. ($\uparrow$,$\downarrow$): higher, lower is better.}
\label{c3-table:human-eval-res}
\end{table}

\section{Conclusions}
We presented an extractive summarization algorithm that controls each summary quality independently, in scenarios where the input is highly redundant.
Redundancy is controlled as the input is consumed, and informativeness and cohesion are balanced during sentence selection.

Results show that our input processing strategy is effective at retrieving non-redundant yet relevant passages, reducing the redundancy levels the rest of the pipeline is exposed to.
In addition, our sentence selector emulates human memory to keep track of cohesive chains while building the summary, enforcing ties between noun phrases directly.
Extensive automatic and human experiments revealed that it is possible to extract highly cohesive summaries that are as informative as summaries optimizing only for informativeness.

\section*{Limitations}

The proposed system presents the following limitations.
First, the system extracts complete sentences and concatenates them to form the final summary.
We do not perform any kind of post-editing of discourse markers that might break coherence in the summary.
However, our results show that the extracted summaries are still perceived as cohesive by humans.
Nevertheless, post-editing is an interesting focus for future work.

Second, we argue about the usefulness of an extractive system in a generative landscape where large language models are predominant.
Recent large language models have shown impressive capabilities at producing coherent, assertive text, some even capable of consuming long sequences of tokens.
However, hallucinations are a pervasive problem in these systems, especially in highly technical domains like the ones considered in this work.
In this scenario, an extractive summary has the advantage of presenting information from the source \textit{verbatim} and hence, without any hallucination.
Moreover, extracted summaries preserve the writing style of the input as well as technical, domain-specific terms,
avoiding altogether the problems of over-simplification and misstyling.

\bibliography{main}
\bibliographystyle{acl_natbib}


\newpage

\appendix

\section{Dataset Preprocessing and Statistics}

For all datasets, we homogenize the source-target length distributions by discarding 
samples with references that were too short (less than 3 sentences, not usefull for our cohesion analysis) or too long (more than 500 tokens in all datasets except \textsc{GovReport}, for which this threshold is set to \num{1000}).
Similarly, samples with short input documents (less than 3 sentences or less than 30 tokens in total) were also discarded.
Sentences were re-split using spaCy\footnote{\url{https://spacy.io/}} and trimmed to 100 tokens, whilst sentences with less than \num{5} tokens were discarded.
Table~\ref{table:data-stats} presents the statistics of all dataset in terms of number of tokens.

It is worth noting that we found a discrepancy in \textsc{PubMed}.
Text from the `\texttt{article}' field (in theory the concatenated sections) would not always have the same text as the `\texttt{sections}' field. Hence, we chose data from the `\texttt{sections}' field as input document.



\paragraph{Datasets with medium-length documents.}
Previous work in redundancy reduction \cite{xiao2020systematically} showed that medium-length documents (e.g.\ articles in CNN-DailyMail, \citealt{nallapati2016abstractive}) do not exhibit content redundancy as acutely as long documents (e.g.\ scientific articles).
In this work, we focus on cases where input redundancy is especially acute.
However, we provide an analysis of the impact of input redundancy on summary properties in Section 6.1. In Figure 2, we observe how summary informativeness and redundancy change as we increase the permissible input length, effectively simulating scenarios with documents of low-, medium-, and high-length.

In addition, the CNN/DM dataset was not included in our experiments because the reference summary is formatted as a list of sentences (highlights) where each sentence exhibits a high degree of independence.
In contrast, a format of continuous sentences such as scientific abstracts constitutes a more appropriate textual format for finding and analyzing cohesion phenomena.

\begin{table}[h]
\centering
\small
\begin{tabular}{lrrrr}
\toprule
\multicolumn{1}{c}{\multirow{2}{*}{\textbf{Dataset}}} & \multicolumn{3}{c}{\textbf{Input Length}}                                                                & \multicolumn{1}{c}{\textbf{Target Len.}} \\
\multicolumn{1}{c}{}                                  & \multicolumn{1}{c}{\textbf{Avg.}} & \multicolumn{1}{c}{\textbf{Max.}} & \multicolumn{1}{c}{\textbf{Q90}} & \multicolumn{1}{c}{\textbf{Avg.}}        \\ \midrule
PubMed                                                & 3150                              & 119875                            & 5844                             & 206                                      \\
BigPatent.C                                           & 4534                              & 72835                             & 8655                             & 119                                      \\
GovReport                                             & 8840                              & 206622                            & 15752                            & 580                                      \\
MultiNews                                             & 2057                              & 525348                            & 3846                             & 260                                      \\ \midrule
CNN/DM                                             & 2057                              & 525348                            & 3846                             & 260                                      \\
NewsRoom                 & 2057                              & 525348                            & 3846                             & 260                                      \\ \bottomrule
\end{tabular}
\caption{Dataset statistics in terms of number of tokens showing average, maximum, and 90\% quantile (Q90).}
\label{table:data-stats}
\end{table}

\section{Optimization and Resource Details}

Long-T5 models were trained using one NVIDA A100 (80Gb of GPU memory).
Table~\ref{table:hyperparams} provides a comprehensive account of hyperparameter values used 
for training and inference in our experiments, for all datasets.
Hyper-parameters specific to selector were optimized w.r.t.\ each dataset's validation set.

The Local Encoder (LE) module is finetuned from pretrained Huggingface's checkpoint 
\texttt{google/long-t5-tglobal-base}, whereas the Global Context Encoder (GCE) is trained from scratch.
For training LE, we obtain extractive oracle sentences from each block and
train the module over blocks with $\text{ROUGE-1}+\text{ROUGE-2}>0.5$.
This threshold was obtained by observing the mean of the distribution of ROUGE-1 + ROUGE-2 scores of these extractive oracles w.r.t. references in preliminary analyses.

During inference,
our system consumes input texts of up to \num{16384} pieces in length.
This input is reduced to a maximum of $M=$\num{1000} sentences, which are pre-extracted by
the LE module and then passed through the GCE module.
From these $M$ sentences, the summary extractor module extracts a summary with a predefined word budget.

\paragraph{Artifact Lincense.}
The models proposed in this paper extend the licenses of the artifacts they are made of.
LongT5 and LED were released under the Apache 2.0 license.
LLaMa is under META's Community License Agreement.\footnote{\url{https://github.com/facebookresearch/llama/blob/main/LICENSE}}

\begin{table}[h]
\centering
\small
\begin{tabular}{p{5.7cm}r}
\toprule
\multicolumn{1}{l}{\textbf{Parameter}} & \multicolumn{1}{c}{\textbf{Value}} \\ \midrule
\textbf{Block Selection}               &                                    \\
Block length in tokens                 & \num{2048}                               \\
Overlapping context size in tokens     & 200                                \\
Damping factor ($d$)                   & 0.85                                \\
Trade-off param. ($\lambda_b$)             & 0.2                                \\ \midrule
\textbf{Local Context Extractor}       &                                    \\
Optimizer                              & Adam                               \\
Learning rate                          & 1E-06                           \\
Learning rate scheduler                & Const.                           \\
Batch size                             & 64                                 \\
Max. gradient norm                     & 2                                  \\
Training steps                         & \num{100000}                             \\
Max. input length in tokens            & \num{2048}                               \\
Max. \# of sentences extracted       & 10                                 \\ \midrule
\textbf{Global Contetext Encoder}      &                                    \\
\# Attention heads                   & 8                                  \\
\# Layers                            & 1                                  \\
Output layer size                      & 200                                \\
Dropout                                & 0.1                                \\
Optimizer                              & Adam                               \\
Learning rate                          & 1E-06                           \\
Learning rate scheduler                & Const.                           \\
Max. input length in tokens            & \num{16384}                              \\
Max. input length in sentences         & \num{1000}                               \\
Batch size                             & 64                                 \\
Max. gradient norm                     & 1                                  \\
Training steps                         & \num{50000}                              \\ \midrule
\textbf{Sentence Selector}      &                                    \\
\textbf{All selectors.}
Trade-off param. ($\lambda_{\text{sel}}$) & 0.8 \\
Summary budget in number of tokens & \\
\hspace{3mm}\textit{PubMed} & 200 \\
\hspace{3mm}\textit{BigPatent.C} & 100 \\
\hspace{3mm}\textit{GovReport} & 650 \\
\hspace{3mm}\textit{MultiNews} & 250 \\
\textbf{KvD-Selector.} & \\
Working memory (\texttt{WM}) & 6 \\
Min. NP cos. similarity ($\nu$) & 0.6 \\
Recall cost ($\gamma_{\text{rec}}$) & 0.01 \\ \bottomrule
\end{tabular}
\caption{Hyper-parameter values for all modules in our summarization pipeline.}
\label{table:hyperparams}
\end{table}


\section{Complementary Results}
\label{appnd:results}
In this appendix, we present additional results in terms of metrics and datasets for analysis in \S6.

\subsection{Additional Metrics}

\paragraph{Semantic Relevance.}
Table~\ref{table:bsc-res} shows BERTScore F$_1$ scores \cite{zhang2019bertscore} with importance weighting (IDF) and with underlying models RoBERTa large \cite{roberta-model} and DeBERTa v2 XLarge-MNLI \cite{he2021deberta}, with HuggingFace checkpoint names \texttt{roberta-large} and \texttt{deberta-xxlarge-mnli}, respectively.
Previous work found that DeBERTa obtained higher correlation with human scores than RoBERTa \footnote{\url{https://github.com/microsoft/DeBERTa}}
showed the same system-level ranking.



\paragraph{Cohesion.}
The additional cohesion metrics considered are Lexical Graph \cite{mesgar-strube-2016-lexical}) and the Noun variants of DiscoScore \cite{zhao2023discoscore}, DS-Focus[NN] and DS-Sent[NN].

Lexical Graph (LexG) computes the adjacency matrix of the sentence graph of a text, where two sentences are connected if they have at least two similar-enough content words, i.e.\ if the cosine similarity between their embeddings is greater than a threshold (zero).
DS-Focus[NN] (DS-Foc) computes the semantic difference between a set of nouns (\textit{foci}) common to a candidate summary and a reference.
Similar to Lexical Graph, DS-Sent[NN] (DS-Sen) computes the semantic similarity between nouns in adjacent sentences, inversely weighted by the distance between sentences.


Table~\ref{table:coh-res-compl} presents results for the additional metrics in all datasets and systems.
In terms of LexG scores, we find that KvD-Select outperforms all systems in all datasets except \textsc{PubMed}, indicating that consecutive sentences in summaries extracted by our system contain more cohesive ties connecting highly semantically similar nouns.
Differences between systems in terms of DS-Foc and DS-Sen scores are much less clearer, with DS-Foc seemingly being more indicative of content coverage and DS-Sen more indicative of cohesion.
Nevertheless, KvD-Select is consistently competitive across datasets, even obtaining the highest DS-Sen score for \textsc{MultiNews}.

Lastly, it is important to note that none of the score differences in Table~\ref{table:coh-res-compl} were found to be statistically significant,
an indicative of the challenge of measuring cohesion.
For this reason, these results were not included in the main argument of this paper.


\begingroup
\setlength{\tabcolsep}{4pt} 

\begin{table*}[t]
\centering
\footnotesize
\begin{tabular}{lcccccccc}
\toprule
\multicolumn{1}{c}{\textbf{System}} & \multicolumn{2}{c}{\textbf{PubMed}} & \multicolumn{2}{c}{\textbf{BigPatent.C}} & \multicolumn{2}{c}{\textbf{GovReport}} & \multicolumn{2}{c}{\textbf{MultiNews}} \\
\textbf{}                           & \textbf{RoB}     & \textbf{DeB}     & \textbf{RoB}        & \textbf{DeB}       & \textbf{RoB}       & \textbf{DeB}      & \textbf{RoB}       & \textbf{DeB}      \\ \midrule
Ext-Oracle                          & 88.44            & 80.20            & 85.83               & 73.80              & 88.30              & 80.06             & 88.69              & 80.04             \\
LT5-Abs                          & \textbf{85.71}            & 73.94            & \textbf{84.09}               & \textbf{70.16}              & \textbf{86.49}              & \textbf{76.52}             & 85.13              & 74.22             \\ \midrule
LED-Ext                            & 83.67            & 69.83            & 83.04               & 68.19              & 85.95              & 74.81             & 85.51              & 73.34             \\
LT5-Ext                         & \textbf{85.71}            & \textbf{74.16}            & 83.77               & 69.81              & 86.44              & 75.95             & \textbf{86.03}              & \textbf{74.34}             \\  \midrule
BlockSel[MemSum]                         & 83.52            & 70.05            & 82.60               & 67.82              & 85.06              & 73.08             & 85.07              & 72.76             \\
BlockSel[LLaMa]                          & 82.86            & 68.52            & 83.17               & 68.65              & 84.80              & 72.82             & 85.33              & 73.03             \\
BlockSel[LT5]                            & 85.05            & 73.08            & 83.65\dag               & 69.75\dag              & 86.46\dag              & 76.06\dag             & 85.97              & 74.10             \\ \midrule
+MMR-Select                         & 85.05            & \red{73.07}\dag            & \blue{83.66}\dag               & \blue{69.78}\dag              & \blue{\textbf{86.49}}              & \blue{76.08}             & \red{85.93}\dag              & \red{74.04}             \\
+NPass                           & \blue{85.13}            & \blue{73.21}            & \blue{83.67}\dag               & \blue{69.76}\dag              & \blue{86.47}\dag              & \blue{76.08}             & \blue{86.01}              & \blue{74.17}\dag             \\
+CCL-Select                         & \red{84.99}\dag            & \red{72.98}            & \red{83.63}               & \red{69.62}              & \blue{86.47}\dag              & \red{76.05}\dag             & \red{85.91}\dag              & \blue{74.18}\dag             \\
+\textbf{KvD-Select}                         & \red{84.76}            & \red{72.43}            & \red{83.34}               & \red{69.15}              & \red{85.99}              & \red{75.17}             & \red{85.72}              & \red{73.67}            \\ \bottomrule
\end{tabular}
\caption{Semantic relevance of system summaries in terms of BERTScore F$_1$ using RoBERTa (RoB) and DeBERTa (DeB) as base models.
\dag: no stat.\ difference between systems in the same column.
Best systems are \textbf{bolded}; systems better than BlockSel[LT5] shown in \blue{blue} and worse, in \red{red}.}
\label{table:bsc-res}
\end{table*}

\endgroup


\begingroup
\setlength{\tabcolsep}{4pt} 

\begin{table*}[t]
\centering
\scriptsize
\begin{tabular}{lcccccccccccc}
\toprule
\multirow{2}{*}{\textbf{Systems}} & \multicolumn{3}{c}{\textbf{PubMed}}                            & \multicolumn{3}{c}{\textbf{BigPatent.C}}                       & \multicolumn{3}{c}{\textbf{GovReport}}  & \multicolumn{3}{c}{\textbf{MultiNews}}                         \\
                                  & \textbf{LGr} & \textbf{DS-Foc} & \textbf{DS-Sen} & \textbf{LGr} & \textbf{DS-Foc} & \textbf{DS-Sen} & \textbf{LGr}      & \textbf{DS-Foc} & \textbf{DS-Sen} & \textbf{LGr} & \textbf{DS-Foc} & \textbf{DS-Sen} \\ \midrule
Gold                              & 1.1088            & 0.0000              & 1.0000               & 0.8368            & 0.0000              & 1.0000               & 1.9990             & 0.0000              & 1.0000               & 0.7756            & 0.0000              & 1.0000               \\
Ext-Oracle                        & 1.1463            & 0.5185              & 0.9692               & 0.8022            & 0.5642              & 0.9499               & 1.9735             & 1.7069              & 0.9973               & 0.7488            & 0.3684              & 0.9789               \\ 
LT5-Abs        & 1.0953                   & 1.2290                     & 0.9262                      & 0.8411                        & 5.0947                          & 0.9106                           & 1.6333                      & 7.9962                        & 0.9896                         & 0.8492                      & 0.7973                        & 0.9602                         \\ \midrule
LT5-Flat                       & \textbf{1.2467}            & 1.0925              & \textbf{0.9350}               & 0.8084            & 1.0244              & \textbf{0.9256}               & 2.0583             & 3.8380              & 0.9949               & 0.9687            & 0.7623              & 0.9642               \\
LED-Flat                          & 1.1634            & 1.1075              & 0.9269               & 0.7800            & 1.0029              & 0.9250               & 1.9735             & 3.7556              & \textbf{0.9953}               & 0.8762            & 0.6672              & 0.9646               \\ \midrule
MemSum-Casc                            & 0.9608            & 0.9232              & 0.9280               & 0.7042            & \textbf{0.9232}              & 0.9174               & 1.8057             & \textbf{3.1476}              & \textbf{0.9953}               & 0.9081            & \textbf{0.6404}              & 0.9630               \\
LlaMa-Casc                             & 0.9222            & \textbf{0.9023}              & 0.9247               & 0.7795            & 0.9315              & 0.9235               & 1.6541             & 3.1698              & 0.9947               & 0.8295            & 0.6774              & 0.9638               \\
BlockSel[LT5]                       & 1.2204            & 1.0219              & 0.9303               & 0.7858            & 1.0291              & 0.9228               & 2.0838             & 3.4428              & 0.9947               & 0.9628            & 0.7353              & 0.9629               \\ \midrule
+MMR-Select                       & 1.2203            & \blue{1.0117}              & 0.9300               & \red{0.7803}            & \blue{1.0108}              & \blue{0.9239}               & \red{2.0797}             & \blue{3.4391}              & 0.9948               & \red{0.9464}            & \blue{0.7348}              & \blue{0.9632}               \\
+NPass                         & \red{1.2153}            & \blue{0.9833}              & 0.9306               & \red{0.7837}            & \blue{1.0246}              & 0.9229               & \red{2.0823}             & \blue{3.4129}              & 0.9946               & \red{0.9595}            & \blue{0.7183}              & 0.9630               \\
+CCL-Select                       & \red{1.2075}            & \blue{0.9970}              & 0.9305               & \red{0.7623}            & \blue{0.9869}              & \blue{0.9233}               & \blue{2.0867}             & \red{3.4481}              & 0.9948               & \red{0.9323}            & \blue{0.7150}              & \blue{0.9638}               \\
+KvD-Select(ours)              & \red{1.1933}            & \blue{0.9948}              & \red{0.9298}               & \blue{\textbf{0.8207}}            & \red{1.0644}              & \red{0.9210}               & \blue{\textbf{2.1824}}             & \red{3.5580}              & 0.9945               & \blue{\textbf{1.0536}}            & \blue{0.7235}              & \blue{\textbf{0.9651}}              \\ \bottomrule
\end{tabular}
\caption{Summary cohesion in terms of Lexical Graph (LexG; \citet{mesgar-strube-2016-lexical}) and DiscoScore's \cite{zhao2023discoscore} DS-Focus[NN] (DS-Foc) and DS-Sent[NN] (DS-Sen).
For all metrics, higher value is better except for DS-Foc.
Best systems are \textbf{bolded}; systems better than BlockSel[LT5] shown in \blue{blue} and worse, in \red{red}.}
\label{table:coh-res-compl}
\end{table*}
\endgroup


\subsection{Additional Comparison Systems and Results}
\label{apx-additional-systems}
We compare against the following additional baselines.

\paragraph{Extractive Oracle.}
The standard extractive oracle, labeled as \textsc{Ext-Oracle}, is obtained by greedily selecting sentences maximizing $\text{ROUGE-1}+\text{ROUGE-2}$ against gold summaries until the word budget is met.

\paragraph{Abstractive Baselines.}
We compare against Longformer Encoder Decoder (LED-Abs;\citet{Beltagy2020Longformer}) and the complete LongT5 encoder-decoder architecture (LT5-Abs;\citet{guo2022longt5}).
Both systems were trained for 3k steps with batch size of \num{128}, and AdamW optimizers.
For \textsc{LT5-Abs}, we used a constant learning rate of 1E-3, using the Huggingface's checkpoints \texttt{google/long-t5-tglobal-base}.
For \textsc{LED-Abs}, a learning rate of 1E-5 was used, with triangular scheduling using 100 steps for warm-up, and using HF's checkpoint \texttt{allenai/led-base-16384}.
For both systems, inference was done using a beam size of 5 and length penalty of $0.5$, $0.5$, $2.0$, and $1.0$ for PubMed, BigPatent.C, GovReport, and MultiNews, respectively.

\paragraph{Choice of Local Encoder.}
We assess the impact of architectural choice for the Local Encoder module in our pipeline by using 
MemSum \cite{gu2022memsum} and LLaMA-7B \cite{touvron2023llama} as encoders.
The systems, dubbed BlockSel[MemSum] and BlockSel[LLaMa], employ a greedy selector.

\paragraph{Results.}
The results on informativeness are presented in Table~\ref{apx-table:inf-res-compl}, whereas redundancy and cohesion are in Table~\ref{apx-table:coh-res}.
The following insights can be drawn.
Using LLaMA as local encoder allows our system to select --greedily-- sentences that have little lexical overlap between them, prompting 
low summary redundancy scores and in turn lowering cohesion scores.
Moreover, the coverage is severely impacted as seen by the low ROUGE scores.
These results might indicate that finetuning a large pretrained model like LLaMA does not necessarily translate to better informativeness,
performing much lower than a smaller model pretrained on the summarization task.
Perhaps unsurprisingly, task-specific, smaller models can be competitive to massive foundation models trained on 1000x more data.

Using MemSum as the local encoder had a similar outcome, although not as severe as when using LLaMa.
Summaries in \citet{gu2022memsum} were obtained in a scenario where only up to 500 sentences were consumed in the order they appear in the document.
In contrast, in our setup, the compared systems consume up to \num{16384} pieces of input text in the order the block selector module retrieves.
The performance gap between the results in \citet{gu2022memsum} and the ones we report can then be explained by input length and ordering conditions.

\begingroup

\setlength{\tabcolsep}{4pt} 

\begin{table*}[ht]
\centering
\small
\begin{tabular}{lllllllllllll}
\toprule
\multicolumn{1}{c}{\multirow{2}{*}{\textbf{System}}} & \multicolumn{3}{c}{\textbf{PubMed}}                                                                 & \multicolumn{3}{c}{\textbf{BigPatent.C}}                                                            & \multicolumn{3}{c}{\textbf{GovReport}}                                                              & \multicolumn{3}{c}{\textbf{MultiNews}}                                                              \\
\multicolumn{1}{c}{}                                 & \multicolumn{1}{c}{\textbf{R1}} & \multicolumn{1}{c}{\textbf{R2}} & \multicolumn{1}{c}{\textbf{RL}} & \multicolumn{1}{c}{\textbf{R1}} & \multicolumn{1}{c}{\textbf{R2}} & \multicolumn{1}{c}{\textbf{RL}} & \multicolumn{1}{c}{\textbf{R1}} & \multicolumn{1}{c}{\textbf{R2}} & \multicolumn{1}{c}{\textbf{RL}} & \multicolumn{1}{c}{\textbf{R1}} & \multicolumn{1}{c}{\textbf{R2}} & \multicolumn{1}{c}{\textbf{RL}} \\ \toprule

Ext-Oracle                                           & 65.10                           & 37.99                           & 60.76                           & 53.85                           & 23.20                           & 46.90                           & 72.66                           & 40.90                           & 69.36                           & 62.66                           & 33.73                           & 57.93                           \\
LED-Abs	& 45.31	& 20.73	& 41.82	& 35.89	& 14.66	& 31.45	& 54.34	& 24.78	& 51.48	& 46.73	& \textbf{18.93}	& \textbf{43.11} \\
LT5-Abs      & \textbf{46.27}              & \textbf{20.92}              & 42.40              & 37.63                   & \textbf{15.67}                   & 32.84                   & 51.72                 & 24.79                 & 49.03                 & 45.72                 & 17.70                 & 41.86                 \\ \midrule
BlockSel[MemSum]                                               & 40.29                           & 14.85                           & 37.09                           & 36.07                           & 10.79                           & 30.97                           & 54.91                           & 19.66                           & 51.75                           & 44.47                           & 15.28                           & 40.32                           \\
BlockSel[LLaMa]                                                & 37.60                           & 11.86                           & 34.51                           & 36.82                           & 11.24                           & 32.00                           & 54.20                           & 19.02                           & 50.90                           & 45.02                           & 15.48                           & 41.00                           \\

BlockSel[LT5]                                                &        46.16                         &       19.74                          &       \textbf{42.49}                          & \textbf{39.57}                           & 13.25                           & \textbf{34.26}                           & \textbf{59.73}                           & \textbf{26.21}                           & \textbf{56.50}                           & \textbf{46.80}                           & 17.21                           & 42.66                           \\ \bottomrule
\end{tabular}
\caption{Informativeness in terms of ROUGE F$_1$ scores (R1, R2, RL), for different choices of local encoder during block selection. All systems employ a greedy selector.
}
\label{apx-table:inf-res-compl}
\end{table*}

\endgroup


\begingroup
\setlength{\tabcolsep}{4pt} 

\begin{table*}[ht]
\centering
\small
\begin{tabular}{lllllllllllll}
\toprule
\multirow{2}{*}{\textbf{Systems}} & \multicolumn{3}{c}{\textbf{PubMed}}                                                                       & \multicolumn{3}{c}{\textbf{BigPatent.C}}                                                                  & \multicolumn{3}{c}{\textbf{GovReport}}                                                                    & \multicolumn{3}{c}{\textbf{MultiNews}}                                                                    \\
                                  & \multicolumn{1}{c}{\textbf{RdRL}} & \multicolumn{1}{c}{\textbf{EGr}} & \multicolumn{1}{c}{\textbf{CCL}} & \multicolumn{1}{c}{\textbf{RdRL}} & \multicolumn{1}{c}{\textbf{EGr}} & \multicolumn{1}{c}{\textbf{CCL}} & \multicolumn{1}{c}{\textbf{RdRL}} & \multicolumn{1}{c}{\textbf{EGr}} & \multicolumn{1}{c}{\textbf{CCL}} & \multicolumn{1}{c}{\textbf{RdRL}} & \multicolumn{1}{c}{\textbf{EGr}} & \multicolumn{1}{c}{\textbf{CCL}} \\ \midrule
Ext-Oracle     & 13.91                & 0.99                     & 0.43                & 14.70                     & 0.68                          & 0.34                     & 14.20                   & 1.92                        & 0.58                   & 10.08                   & 0.68                        & 0.51                   \\
LED-Abs     & 15.00                    & 0.93                & 0.85                      & 38.23                         & \textbf{0.94}                     & 0.77                    & 16.55                       & 1.75                   & 0.75                    & \textbf{8.37}                        & 0.53                   & \textbf{0.83} \\
LT5-Abs        & 16.15                & 0.96                     & \textbf{0.86}                & 38.04                     & 0.82                          & \textbf{0.81}                     & 15.89                   & 1.60                        & \textbf{0.81}                   & 12.60                   & 0.78                        & 0.78                   \\ \midrule 
BlockSel[MemSum]                            & 12.58                             & 0.75                               & 0.25                             & 19.41                             & 0.62                               & 0.34                             & 13.77                             & 1.72                               & 0.29\dag                             & 12.29\dag                             & 0.84                               & 0.26\dag                             \\
BlockSel[LLaMA]                             & \textbf{11.61}                             & 0.70                               & 0.27                             & \textbf{17.51}                             & 0.70                               & 0.39                             & \textbf{12.43}                             & 1.58                               & 0.28\dag                             & 10.87                             & 0.77                               & 0.25                             \\
BlockSel[LT5]                       & 17.08                    & \textbf{1.07}                               & 0.26                             & 20.15\dag                             & 0.73                               & 0.39                             & 16.34                             & \textbf{2.04}                               & 0.27                             & 12.26\dag                             & \textbf{0.90}                               & 0.26\dag                             \\  \midrule
Gold                              & 13.54                             & 0.95                               & 0.78                             & 18.11                             & 0.78                               & 0.83                             & 13.37                             & 1.95                               & 0.75                             & 9.72                             & 0.71                               & 0.80                             \\ 
\bottomrule
\end{tabular}
\caption{Summary redundancy (RdRL), cohesion (EGr), and local cohesion (CCL), for complementary abstractive, flat, and cascaded systems.
For cohesion and coherence metrics, higher is better.
Best systems are \textbf{bolded}.
\dag: no stat.\ difference between systems in the same column.
}
\label{apx-table:coh-res}
\end{table*}
\endgroup


\subsection{Reducing redundancy in block selection}
\label{apx-brank-all}

Figure~\ref{fig:brank-all} presents the effect of block selection strategies for \textsc{PubMed}, \textsc{BigPatent.C}, and \textsc{GovReport}.

\begin{figure*}[t]
     \centering
     \begin{subfigure}[t]{\textwidth}
         \centering
         \includegraphics[width=\textwidth]{figures/pubmed_brank_rl.pdf}
         \caption{PubMed}
     \end{subfigure}
     \\ \vspace{1em}
     \begin{subfigure}[b]{\textwidth}
         \centering
         \includegraphics[width=\textwidth]{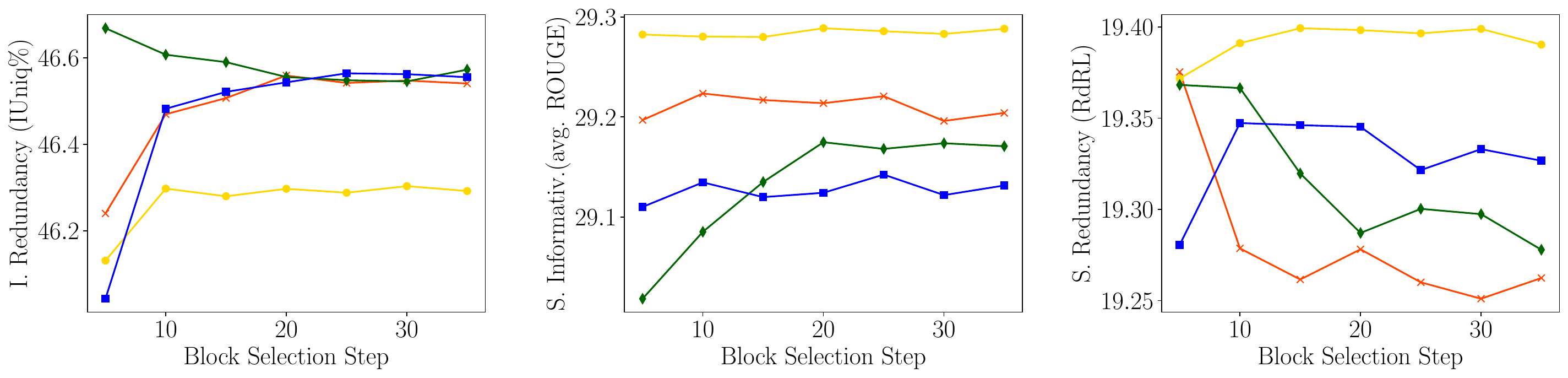}
         \caption{BigPatent.C}
     \end{subfigure}
     \\ \vspace{1em}
     \begin{subfigure}[b]{\textwidth}
         \centering
         \includegraphics[width=\textwidth]{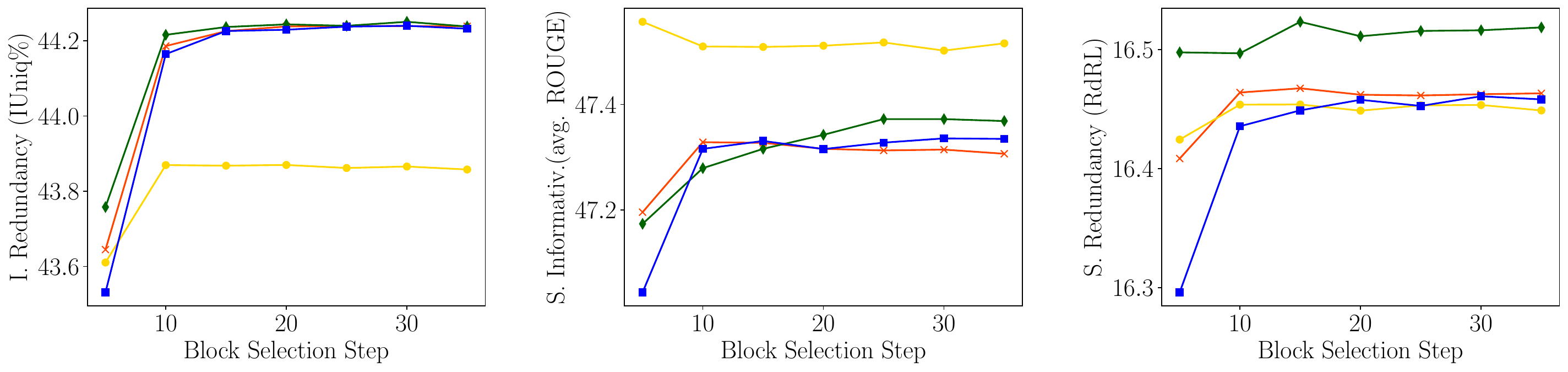}
         \caption{GovReport}
     \end{subfigure}
     \begin{subfigure}[b]{\textwidth}
         \centering
         \includegraphics[width=\textwidth]{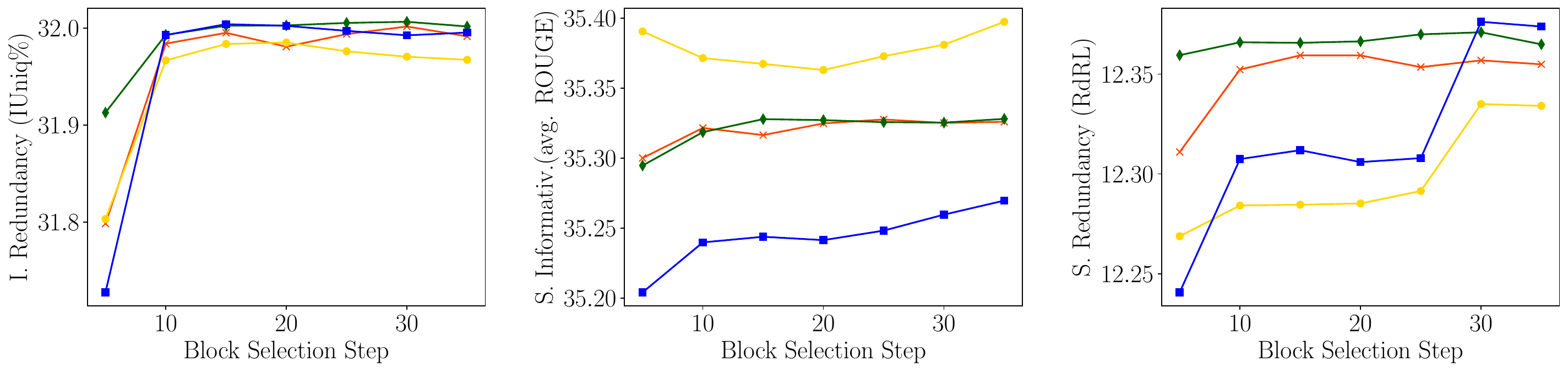}
         \caption{MultiNews}
     \end{subfigure}
     \caption{Effect of block selection strategy over input redundancy (left), summary informativeness (center), and summary redundancy (right), evaluated as block selection proceeds on the validation splits of all datasets analysed.}
    \label{fig:brank-all} 
\end{figure*}

\subsection{Effect of Trade-off Parameter $\lambda_\text{sel}$}
\label{apx-lmb-sel-all}
Figure~\ref{fig:lmb_sel-all} showcases how summary properties (informativeness, redundancy, and cohesion) 
vary across increasing levels of $\lambda_\text{sel}$,
for all datasets analyzed.


\begin{figure*}[t]
     \centering
     \begin{subfigure}[t]{\textwidth}
         \centering
         \includegraphics[width=0.9\textwidth]{figures/pubmed_rl_selectors.pdf}
         \caption{PubMed}
     \end{subfigure}
     \\ \vspace{1em}
     \begin{subfigure}[b]{\textwidth}
         \centering
         \includegraphics[width=0.9\textwidth]{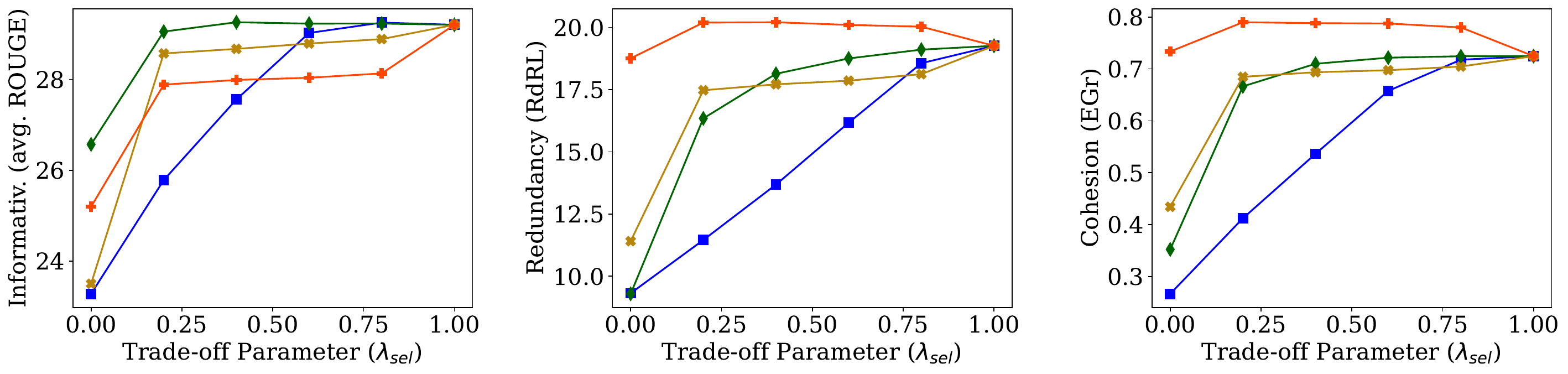}
         \caption{BigPatent.C}
     \end{subfigure}
     \\ \vspace{1em}
     \begin{subfigure}[b]{\textwidth}
         \centering
         \includegraphics[width=0.9\textwidth]{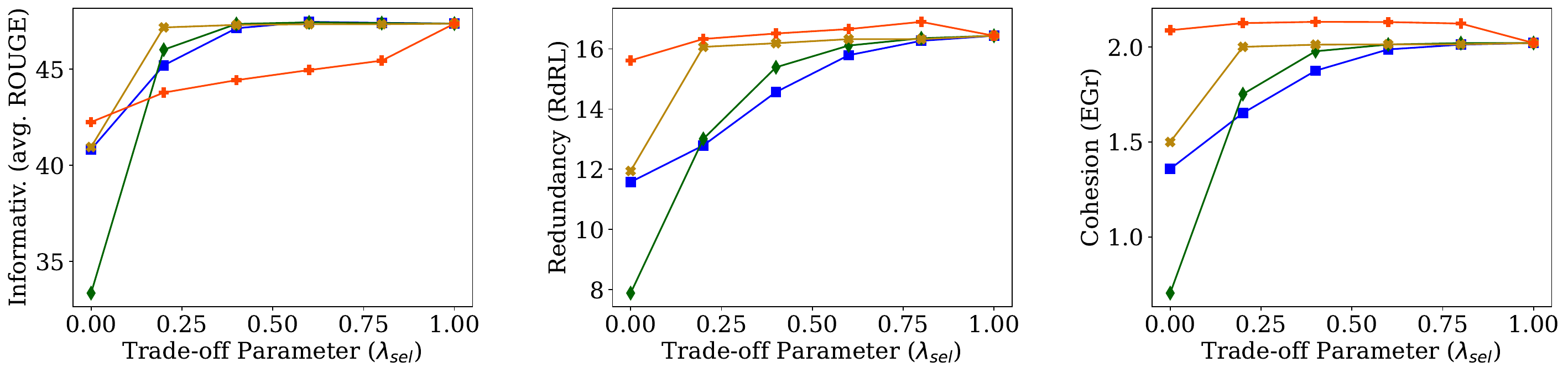}
         \caption{GovReport}
     \end{subfigure}
     \\ \vspace{1em}
     \begin{subfigure}[b]{\textwidth}
         \centering
         \includegraphics[width=0.9\textwidth]{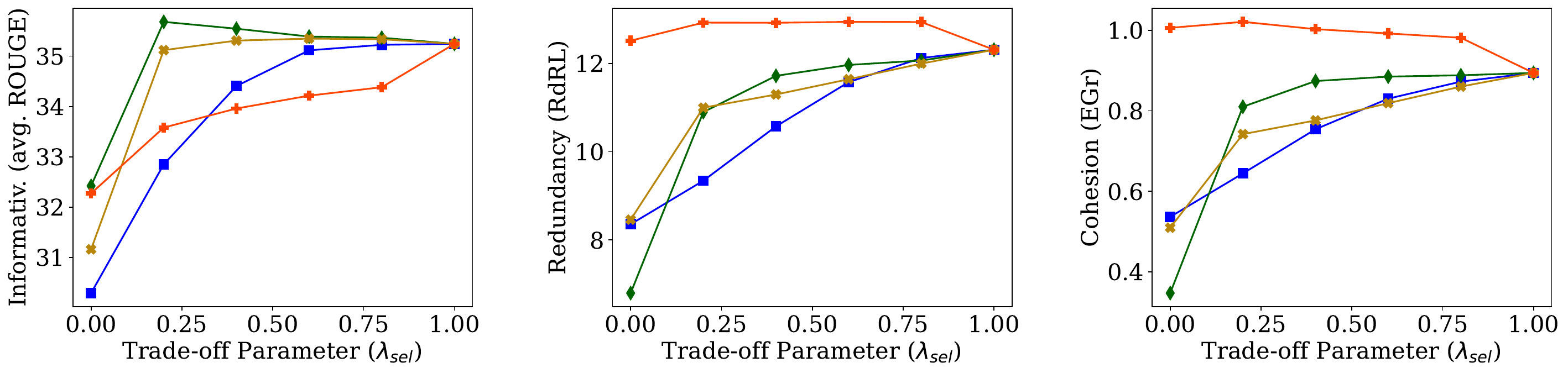}
         \caption{MultiNews}
     \end{subfigure}
    \caption{Informativeness (left), redundancy (mid), and lexical cohesion (right) across different values of the trade-off parameter $\lambda_{sel}$ on the validation set of \textsc{PubMed}, \textsc{BigPatent.C}, \textsc{GovReport}, and \textsc{MultiNews}.}
    \label{fig:lmb_sel-all}
\end{figure*}

\section{Human Evaluation Campaigns}

In this section, we provide further details about the two evaluation campaigns.
Both campaigns were run on Amazon Mechanical Turk, where Turkers were required to have a Human Intelligence Task (HIT) approval rate higher than $99\%$, a minimum of \num{10000} approved HITs, be proficient in the English language, and have worked in the healthcare or medical sector before.

Regarding the difficulty of the domain for non-expert annotators, previous work \cite{fabbri2021summeval} showed that preference rankings of systems were the same across expert and non-expert annotators,
although with more marked differences in preference for expert annotations.
This indicates that non-expert annotation still holds statistical value and conclusions can be drawn from them, as has been done in previous work for arXiv \cite{cho2022toward} and PubMed \cite{gu2022memsum}.

We requested three different annotators per HIT, and annotators were awarded $\$1$ per HIT, translating to more than
\$15 per hour. These rates were calculated by measuring the average annotation time per HIT in a pilot study.
Furthermore, we implemented the following catch controls: 
(i) we asked participants to check checkboxes confirming they had read the instructions and examples provided,
and (ii) we discard HITs that were annotated in less than \num{5} minutes.\footnote{Time threshold obtained from pilot study measurements.}
Annotations that failed the controls were discarded in order to maximize the quality.
Figure~\ref{fig:camp-ins} depicts the instructions given to annotators for each campaign.

\subsection{Ranking Campaign}
We collected three annotations (by different annotators) per system-pair comparison and made sure that the same annotator was not exposed to the same document twice.
As an additional catch trial, we included in each annotation batch an extra instance with summaries extracted by the extractive oracle and the random baseline.

After discarding annotations that failed the controls, we are left with \num{708} out of \num{810} instances (\num{30} documents, \num{3} system pairs, \num{3} dimensions, and \num{3} annotations per pair), and \num{90} different annotators.

\subsection{Chaining Campaign}
Participants were shown a single system summary as a list of sentences where tokens that belonged to the same noun phrase were colored the same.
Then, the task consists of selecting sets of colored text chunks that belong to the same semantic field.
Similarly to the previous study, we collected three annotations per system summary (by different annotators)
and included the gold summary of an extra system in the campaign.

We collected \num{908} human annotations of noun-phrase chains for 360 summaries (\num{30} documents, \num{4} systems including gold summaries, and \num{3} annotations per summary), from a total of \num{86} different annotators.
On average, annotators identified $2.56$ groups per summary and $3.49$ NPs per group.


\begin{figure*}[t]
     \centering
     \begin{subfigure}[b]{\textwidth}
         \centering
         \includegraphics[width=0.8\textwidth]{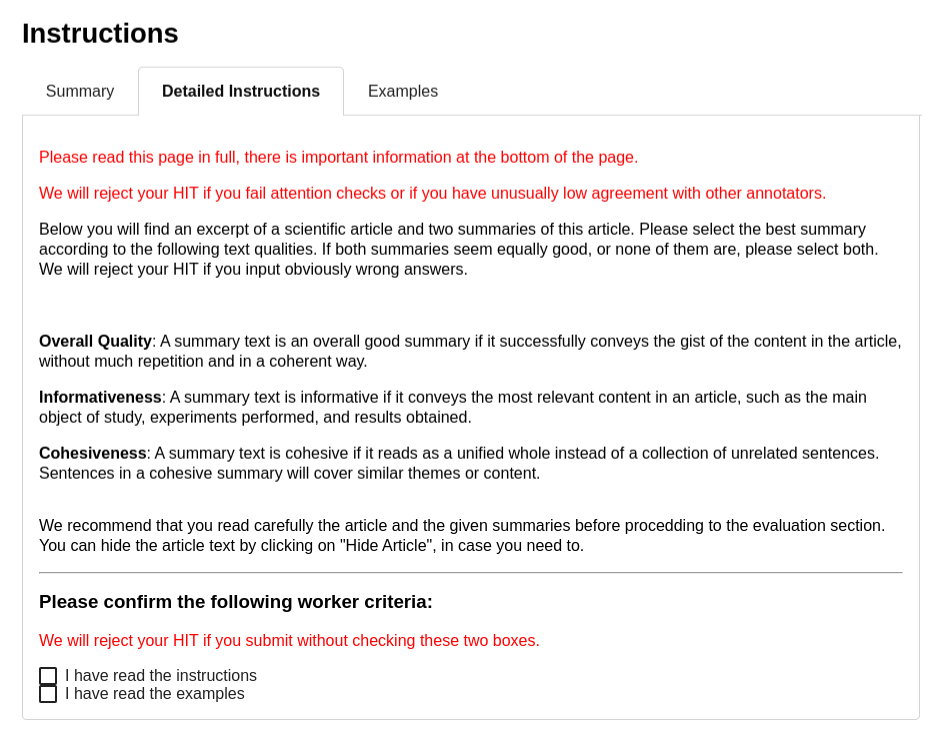}
         \caption{Ranking Campaign}
     \end{subfigure}
     \\ \vspace{1em}
     \begin{subfigure}[b]{\textwidth}
         \centering
         \includegraphics[width=0.8\textwidth]{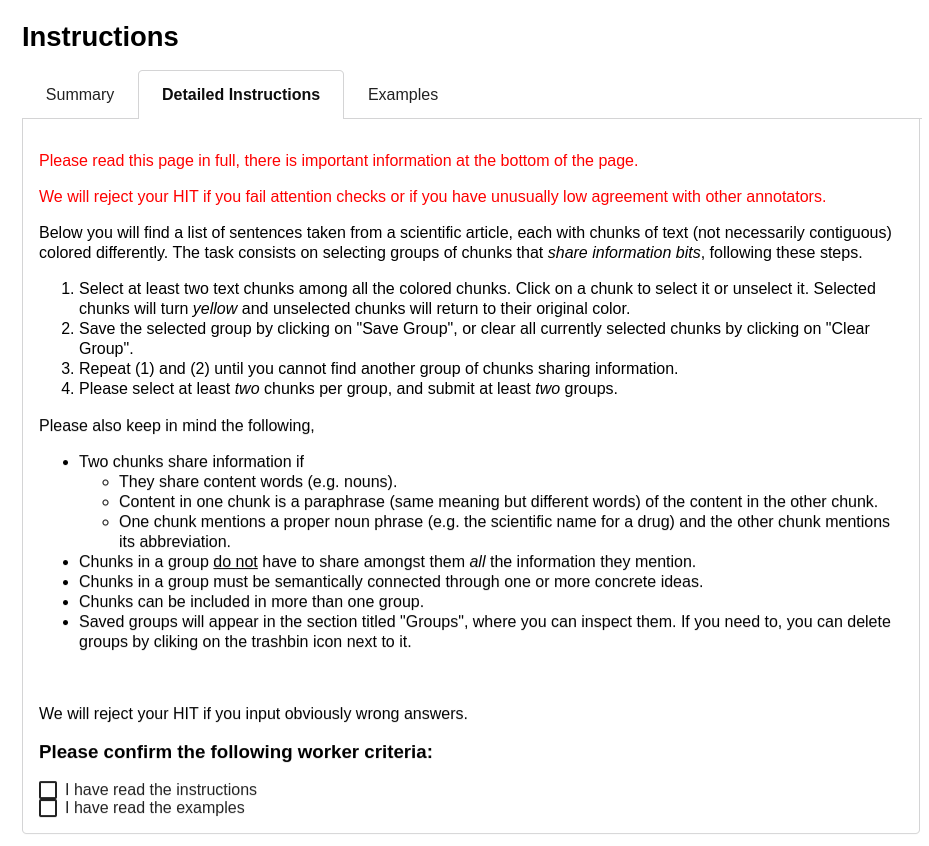}
         \caption{Chaining Campaign}
     \end{subfigure}
    \caption{Instructions given to annotators in the ranking (top) and chaining campaigns (bottom).}
    \label{fig:camp-ins}
\end{figure*}


\section{Example Output}

\begingroup
\setlength{\tabcolsep}{0.6pt}
\renewcommand{\arraystretch}{0.9} 

\begin{table*}[t]
\centering
\scriptsize
\begin{tabular}{p{14.5cm}r}
\toprule
\textbf{System Summary}                                                                                                                                                                                                                                                                                                                                               & \textbf{Chain IDs} \\ \midrule
\textbf{Gold (Avg. ROUGE=-; RdRL=8.3; EGr=0.72)}     &                    \\
Why did \blue{Microsoft} buy \blue{Nokia}'s \textcolor{BrickRed}{phone business}?     & \blue{1},\textcolor{BrickRed}{2}                \\
We now know \blue{Microsoft}'s answer: \blue{the computing giant} released a \textcolor{gray}{30-slide presentation} today arguing that the move will improve \blue{Microsoft}'s margins on \textcolor{BrickRed}{Windows phones}, which will allow it to invest more in the \blue{platform}, which will accelerate sales and market share growth, \textcolor{orange}{The Washington Post} reports.                                                            & \blue{1},\textcolor{BrickRed}{2},\textcolor{orange}{3},\textcolor{gray}{5}            \\
But \textcolor{orange}{John Herrman at BuzzFeed} has another explanation: "fear of dying alone."                                                                                                                                                                                                                                                                                          & \textcolor{orange}{3}                  \\
Here's what \textcolor{orange}{he and other pundits} are saying: \textcolor{gray}{the presentation} "manages to sound both insane and uninspiring, outlining modest goals that still sound unrealistic," \textcolor{orange}{Herrman} argues - like capturing a whole 15\% of the smartphone market.                                                                                                                             & \textcolor{BrickRed}{2},\textcolor{orange}{3},\textcolor{gray}{5}              \\
"It's a fitting end for the close of \blue{Microsoft}'s \textcolor{Cyan}{Ballmer} era, during which the \blue{company}...missed out on the most important change in \textcolor{BrickRed}{consumer electronics} in decades" while remaining profitable in unglamorous ways.                                                                                                                                                  & \blue{1},\textcolor{BrickRed}{2},\textcolor{Cyan}{4}              \\
Like everyone, \blue{Microsoft} is trying to ape the \blue{Apple} model, \textcolor{orange}{MobileOpportunity} observes.                                                                                                                                                                                                                                                                                & \blue{1},\textcolor{orange}{3}                \\
But \textcolor{orange}{it}'s not so sure that's a good idea.                                                                                                                                                                                                                                                                                                                              & \textcolor{orange}{3}                  \\
"There already is an \blue{Apple}," \textcolor{orange}{the blog} points out, and other software/hardware hybrid companies, like \blue{Palm} and \blue{BlackBerry}, have been crushed under its heel.                                                                                                                                                                                                           & \blue{1},\textcolor{orange}{3}                \\
Maybe \blue{Microsoft} should have tried to patch up its tried-and-true strategy of licensing its \textcolor{BrickRed}{OS}.                                                                                                                                                                                                                                                                        & \blue{1},\textcolor{BrickRed}{2}                \\
The move risks complicating \blue{Microsoft}'s crucial relationships with \textcolor{BrickRed}{other PC and device manufacturers}, one analyst tells \textcolor{orange}{ZDNet}.                                                                                                                                                                                                                                        & \blue{1},\textcolor{BrickRed}{2},\textcolor{orange}{3}              \\
But he adds that "\blue{Microsoft} needed to make a bold move" or face "certain terminal decline," and that the price it paid for \blue{Nokia} "seems extremely reasonable."                                                                                                                                                                                                        & \blue{1},\textcolor{orange}{3}                \\
Meanwhile, \textcolor{orange}{Matthew Yglesias at Slate} digs up a fairly interesting memo from \blue{Nokia} \textcolor{Cyan}{CEO} (and, perhaps, \blue{Microsoft} heir apparent) \textcolor{Cyan}{Stephen Elop}, in which he uses the story of a Deepwater Horizon worker leaping from the burning oil platform - a seemingly desperate, yet necessary \textcolor{teal}{move} - to explain \textcolor{teal}{the company's shift} from its own failed \textcolor{BrickRed}{OS} to \textcolor{BrickRed}{Windows Phone}.      & \blue{1},\textcolor{BrickRed}{2},\textcolor{orange}{3},\textcolor{Cyan}{4},\textcolor{teal}{11}         \\
Of course, \textcolor{orange}{Yglesias} notes, \textcolor{teal}{that move} "was basically a total failure."                                                                                                                                                                                                                                                                                                 & \textcolor{orange}{3},\textcolor{teal}{11}               \\ \midrule
\textbf{MMR-Select (Avg. ROUGE=28.25; RdRL=8.31; EGr=0.79)}                                                                                                                                                                                                                                                                                                         &                    \\
Summary: \blue{Microsoft}'s \textcolor{purple}{acquisition} of \blue{Nokia} is aimed at building a \textcolor{BrickRed}{devices and services strategy}, but the \blue{joint company} won't take the same form as \blue{Apple}.                                                                                                                                                                                                              & \blue{1},\textcolor{BrickRed}{2},\textcolor{purple}{10}             \\
\textcolor{MidnightBlue}{This crawl} was run at level 1 (URLs, including their embeds, plus the URLs of all outbound links, including their embeds).                                                                                                                                                                                                                                            & \textcolor{MidnightBlue}{6}                  \\
Today's sale price, which includes \textcolor{Violet}{1.65 billion euros} in \textcolor{BrickRed}{patents}, is just \textcolor{Violet}{5.44 billion euros}.                                                                                                                                                                                                                                                                         & \textcolor{BrickRed}{2},\textcolor{Violet}{7}                \\
It's been a rough decade.                                                                                                                                                                                                                                                                                                                                             &  -                  \\
\blue{Microsoft} is buying \blue{Nokia}'s \textcolor{BrickRed}{cell phone business} and licensing its \textcolor{BrickRed}{patent portfolio}, according to \blue{both companies}.                                                                                                                                                                                                                                                      & \blue{1},\textcolor{BrickRed}{2}                \\
In 2003, \blue{Nokia}'s \textcolor{BrickRed}{cell phone market share} exceeded 35\%.                                                                                                                                                                                                                                                                                                               & \blue{1},\textcolor{BrickRed}{2}                \\
That same year, its \textcolor{BrickRed}{phone business} alone posted an operating profit of \textcolor{Violet}{5.48 billion euros}.                                                                                                                                                                                                                                                                            & \textcolor{BrickRed}{2},\textcolor{Violet}{7}                \\
\blue{Nokia} lashed itself to \blue{Microsoft}'s mast after losing out to \textcolor{BrickRed}{iOS} and \textcolor{BrickRed}{Android} in \textcolor{BrickRed}{the smartphone market share stakes} and \textcolor{Sepia}{with the limited success} of the \textcolor{BrickRed}{Lumia} range so far, enough to keep interest in \textcolor{BrickRed}{Windows Phone} alive, most analysts are seeing a certain amount of inevitability to the \textcolor{purple}{acquisition}, even if they are split on what its biggest implications are. & \blue{1},\textcolor{BrickRed}{2},\textcolor{Sepia}{8},\textcolor{purple}{10}             \\
The seed for \textcolor{MidnightBlue}{this crawl} was a list of every host in the Wayback Machine.                                                                                                                                                                                                                                                                                              & \textcolor{MidnightBlue}{6}                  \\
The WARC files associated with \textcolor{MidnightBlue}{this crawl} are not currently available to the general public.                                                                                                                                                                                                                                                                          & \textcolor{MidnightBlue}{6}                  \\
Five years ago was the year the \textcolor{BrickRed}{App Store} first opened.                                                                                                                                                                                                                                                                                                               & \textcolor{BrickRed}{2}                 \\
\textcolor{BrickRed}{Windows Phone} has barely dented the now much larger \textcolor{BrickRed}{smartphone market}.                                                                                                                                                                                                                                                                                                & \textcolor{BrickRed}{2}                 \\
Many at the time wondered if \textcolor{Cyan}{Stephen Elop}'s time at \blue{Nokia} would be spent grooming the \blue{company} for purchase —a foreigner in all possible ways, he began his time at the \blue{company} with a memo rightly but offensively declaring \blue{Nokia}'s proud platform a failure and quickly pledged the \blue{company}'s commitment to the still-tiny \textcolor{BrickRed}{Windows Phone}.                            & \blue{1},\textcolor{BrickRed}{2},\textcolor{Cyan}{4}                \\ \midrule
\textbf{KvD-Select (Avg. ROUGE=26.33; RdRL=12.48; EGr=0.88)}                                                                                                                                                                                                                                                                                                        &                    \\
Summary: \blue{Microsoft}'s \textcolor{purple}{acquisition} of \blue{Nokia} is aimed at building a \textcolor{BrickRed}{devices and services strategy}, but \blue{the joint company} won't take the same form as \blue{Apple}.                                                                                                                                                                                                              & \blue{1},\textcolor{BrickRed}{2},\textcolor{purple}{10}             \\
\blue{Microsoft} has been working on its evolution into a \textcolor{BrickRed}{devices and services company}, moving away from the \textcolor{BrickRed}{services business} it has traditionally been, for several years now \textcolor{Sepia}{with limited success}.                                                                                                                                                                        & \blue{1},\textcolor{BrickRed}{2},\textcolor{Sepia}{8}              \\
\blue{Nokia} lashed itself to \blue{Microsoft}'s mast after losing out to \textcolor{BrickRed}{iOS} and \textcolor{BrickRed}{Android} in \textcolor{BrickRed}{the smartphone market share stakes} and \textcolor{Sepia}{with the limited success} of the \textcolor{BrickRed}{Lumia} range so far, enough to keep interest in \textcolor{BrickRed}{Windows Phone} alive, most analysts are seeing a certain amount of inevitability to the \textcolor{purple}{acquisition}, even if they are split on what its biggest implications are. & \blue{1},\textcolor{BrickRed}{2},\textcolor{Sepia}{8},\textcolor{purple}{10}             \\
Owning \textcolor{BrickRed}{the desktop (via Windows)} and building additional services on top, like \textcolor{BrickRed}{Office} or \textcolor{BrickRed}{Search}, has been vital for \blue{Microsoft}'s strategy until now, so, as our interest shifts from the \textcolor{BrickRed}{desktop} to the \textcolor{BrickRed}{tablet} or \textcolor{BrickRed}{smartphone}, it's essential to \blue{Microsoft}'s \textcolor{BrickRed}{broader business (even Azure)} that it can \textcolor{OliveGreen}{retain that connection in some form}.                             & \blue{1},\textcolor{BrickRed}{2},\textcolor{OliveGreen}{9}              \\
But he said \blue{Microsoft}'s challenge remains \textcolor{OliveGreen}{how to unite the myriad services and brands} - \textcolor{BrickRed}{Windows}, \blue{Nokia}, \textcolor{BrickRed}{Live}, \textcolor{BrickRed}{Surface}, \textcolor{BrickRed}{Xbox}, \textcolor{BrickRed}{Bing}, and more - into a cohesive experience that will command and cement customer loyalty.                                                                                                                                               & \blue{1},\textcolor{BrickRed}{2},\textcolor{OliveGreen}{9}                \\
It felt like a radical about-face, but no matter: \blue{Nokia} and \blue{Microsoft} were going to \textcolor{OliveGreen}{save each other}.                                                                                                                                                                                                                                                                  & \blue{1},\textcolor{OliveGreen}{9}                \\ \bottomrule

\end{tabular}
\caption{Reference summary, along with summaries extracted by \textsc{MMR-Select} and \textsc{KvD-Select} for a \textsc{MultiNews} sample with informativeness (average ROUGE score), redundancy (RdRL), and cohesion (EGr) scores.
Each sentence is annotated with lexical chains, color-coded in the text and IDs shown to the right.
Text was detokenized and truecased for ease of reading. }
\label{c3-table:qualtv}
\end{table*}
\endgroup

\end{document}